\documentclass{svproc}
\usepackage{url}
\usepackage{graphicx}%
\usepackage{multirow}%
\usepackage{amsmath,amssymb,amsfonts}%
\usepackage{rotating}
\usepackage{mathrsfs}%
\usepackage[title]{appendix}%
\usepackage{xcolor}%
\usepackage{textcomp}%
\usepackage{manyfoot}%
\usepackage{booktabs}%
\usepackage{algorithm}%
\usepackage{algorithmicx}%
\usepackage{algpseudocode}%
\usepackage{listings}%
\usepackage{comment}

\begin{document}
\mainmatter              
\title{Deep learning-based method for weather forecasting: A case study in Itoshima}
\titlerunning{Deep learning-based method for weather forecasting}  
%
\author{Yuzhong Cheng\inst{1} \and Linh Thi Hoai Nguyen\inst{2} \and
Akinori Ozaki\inst{3}   \and Ton Viet Ta\inst{1,4} }
\authorrunning{Yuzhong Cheng et al.} 
%
\tocauthor{Yuzhong Cheng, Linh Thi Hoai Nguyen, 
Akinori Ozaki  and Ton Viet Ta}
\institute{Joint Graduate School of Mathematics for Innovation, Kyushu University\\
\email{cheng.yuzhong.129@s.kyushu-u.ac.jp}\\ 
\and
International Institute for Carbon-Neutral Energy Research, Kyushu University\\
\email{linh@i2cner.kyushu-u.ac.jp}
\and
Institute of Tropical Agriculture, Kyushu University\\
\email{a-ozaki@agr.kyushu-u.ac.jp}
\and
Faculty of Agriculture, Kyushu University\\
\email{tavietton@agr.kyushu-u.ac.jp}\\
744 Motooka, Nishi-ku, 819-0395 Fukuoka, Japan}

\maketitle              

\begin{abstract}
Accurate weather forecasting is of paramount importance for a wide range of practical applications, drawing substantial scientific and societal interest. However, the intricacies of weather systems pose substantial challenges to accurate predictions. This research introduces a multilayer perceptron model tailored for weather forecasting in Itoshima, Kyushu, Japan. Our meticulously designed architecture demonstrates superior performance compared to existing models, surpassing benchmarks such as Long Short-Term Memory and Recurrent Neural Networks. 
\end{abstract}
\allowdisplaybreaks
\section{Introduction}

Weather forecasting is an age-old challenge that has consistently captivated researchers due to its multifaceted implications across various domains, including agriculture, ecology, and political decision-making. While numerous studies have explored weather forecasting in a broader context, the complexity of weather systems demands a focus on specific regions to yield meaningful insights.

In this study, we direct our attention to Itoshima, a city nestled in Fukuoka Prefecture, Japan. Itoshima's heavy reliance on agriculture and tourism renders it particularly susceptible to the vagaries of weather conditions. The accuracy of weather forecasting holds significant economic implications for the city, making it imperative to enhance prediction methodologies. In response to this demand, we delve into the realm of deep learning techniques, leveraging their potential to revolutionize weather forecasting accuracy in this unique geographical setting.

Traditionally, weather forecasting entailed the analysis of time series data, encompassing parameters such as temperature and rainfall. Historical weather data provided the foundation for predicting future conditions, a task often approached through statistical or numerical methods. Although models like regression and stochastic processes found application in time series data analysis, they frequently struggled to capture the intricate dynamics of weather conditions (\cite{kothapalli2017real,tektacs2010weather}).

Numerous studies have explored the application of big data tools to weather forecasting, yet limitations persist, particularly concerning data structure and optimization algorithms (\cite{fathi2022big}). This limitation underscores the need for innovative approaches.

Deep neural networks, with their multilayer architecture, offer a potent solution for capturing the intricate, nonlinear relationships inherent in weather data. Depending on data characteristics, various types of deep neural networks, such as Multilayer Perceptron (MLP) and Recurrent Neural Network (RNN)-based models, have emerged as effective choices for weather forecasting (\cite{narvekar2015daily,salman2015weather,barrera2022rainfall,ren2021deep}). These models have demonstrated their prowess in enhancing prediction precision over traditional methods.

This study represents a dedicated effort to enhance the precision of weather forecasting in the Itoshima region by constructing novel deep neural network. We introduce a carefully constructed Multilayer Perceptron (MLP) model, trained on an extensive dataset collected by our laboratory in Itoshima. The dataset includes seven critical weather condition variables (temperature, solar radiation, wind speed, wind direction, relative humidity, rainfall, and atmospheric pressure), tailored for optimal use with the MLP model. Our comprehensive analysis of the proposed model highlights its exceptional performance in accurately estimating key variables. Importantly, our model serves as a pioneering innovation, presenting the first simultaneous forecasting of all the seven variables specific to the Itoshima area.

The subsequent sections of this paper are structured as follows: Section 2 introduces the dataset utilized in this study and provides an overview of the methodology employed for weather forecasting using the MLP model. Section 3 offers insights into the results obtained during the training process and assesses prediction performance on the test set. Finally, in Section 4, we draw overarching conclusions about our method's efficacy and suggest avenues for future research.

\section{Materials and Method}

In this section, we present a comprehensive description of the dataset used in our study and introduce the neural network model that forms the core of our analysis. We delve into the dataset's key features, detailing its composition and relevance, and provide an in-depth exploration of the network's architecture, including both its structural components and the training algorithm we employed.

\subsection{Materials}

\subsubsection{Data description}
\begin{figure}[H]
	\centering
	\includegraphics[scale=0.27]{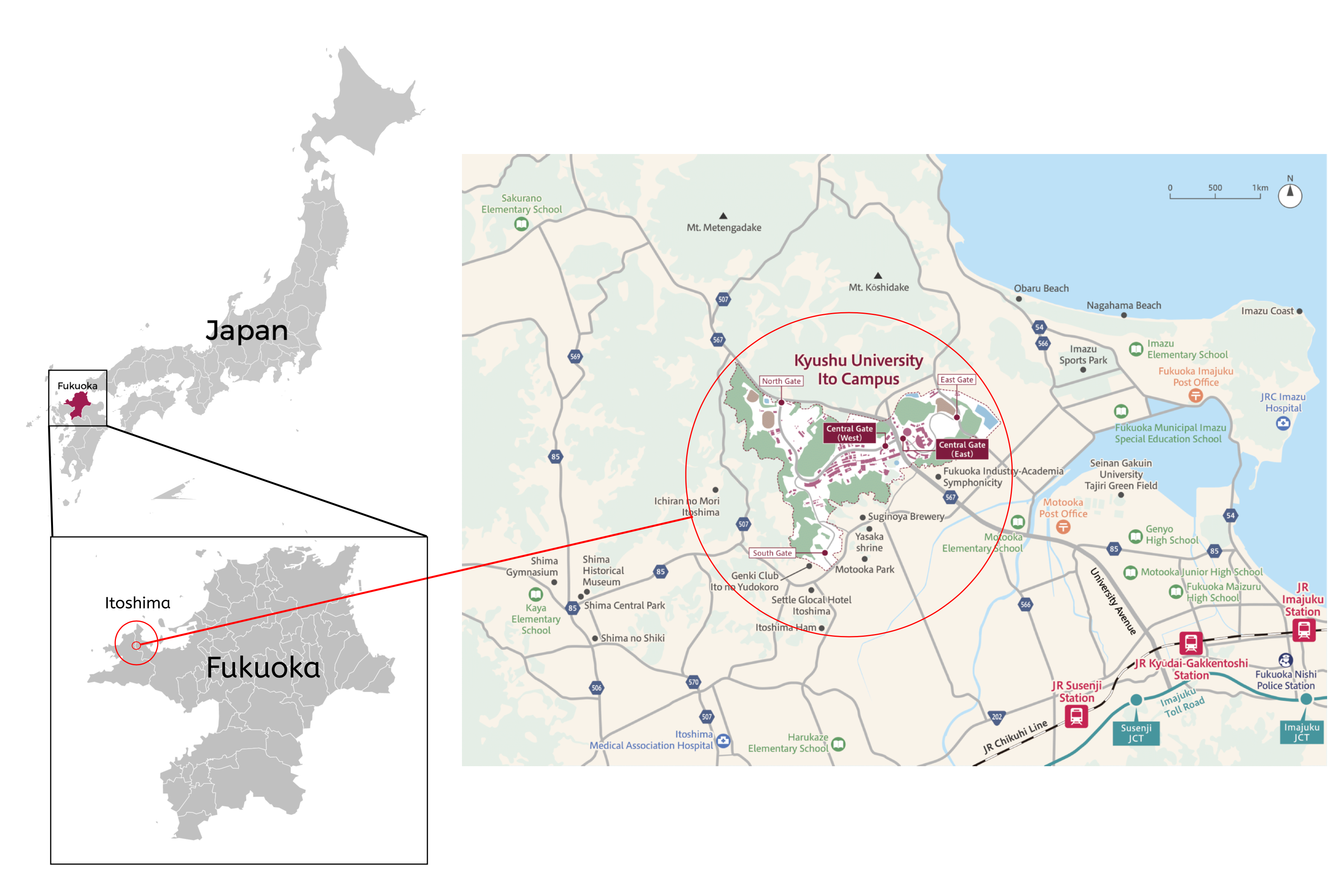}
	\caption{Location of Ito campus, Kyushu University}
	\label{Kyudai}
\end{figure}

Our study focuses on the geographical region of Itoshima, 
a coastal city located in northwestern Fukuoka Prefecture, Japan, with approximate coordinates between 33.5000° N and 33.7500° N latitude and 130.2500° E and 130.7500° E longitude. Itoshima is characterized by a diverse agricultural landscape, attributed to its unique geographical setting that encompasses coastal areas, rolling hills, and fertile plains."
We collected our data from a weather station situated in the Agriculture zone at Ito campus, Kyushu University, within the Itoshima area, see Figure \ref{Kyudai}.
The data collection sensors used were manufactured by were manufactured by CLIMATEC, Inc. and included a wind velocity and direction sensor (CYG-5108), an ambient temperature and humidity sensor (CVS-HMP155D), a solar radiation sensor (CHF-SR20), a rainfall sensor (CTKF-1), a pressure sensor (CVS-PTB110), and a datalogger (CR1000 from Campbell Scientific).

Our study utilizes a comprehensive dataset that meticulously records various weather parameters in the Itoshima region.  This dataset covers a time span of more than a month in 2021, commencing on October 15, 2021, and concluding on December 1, 2022. Furthermore, it spans over 7 months in 2022, beginning on April 1, 2022, and ending on November 1, 2022. Observations were recorded at 10-minute intervals, resulting in a total of 37,585 data points, with 6,769 observations recorded in 2021 and 30,816 in 2022.

The weather station we employed for data collection yielded highly accurate readings of seven essential weather parameters: temperature (recorded in Celsius), solar radiation (measured in $W/m^2$), wind speed (expressed in $m/s$), wind direction (reported in degrees), relative humidity (indicated in percentage), rainfall (quantified in $mm/10min$), and atmospheric pressure (logged in $hPa$).

These seven parameters, meticulously collected and analyzed in our dataset, hold significant importance in comprehending weather patterns and forecasting future weather conditions. They provide invaluable insights for assessing the potential impacts of weather on ecological and agricultural processes, including aspects like plant growth, animal behavior, and crop yield.

\begin{figure}[H]
\begin{minipage}{0.3\textwidth}
		\centering
		\includegraphics[width=\textwidth]{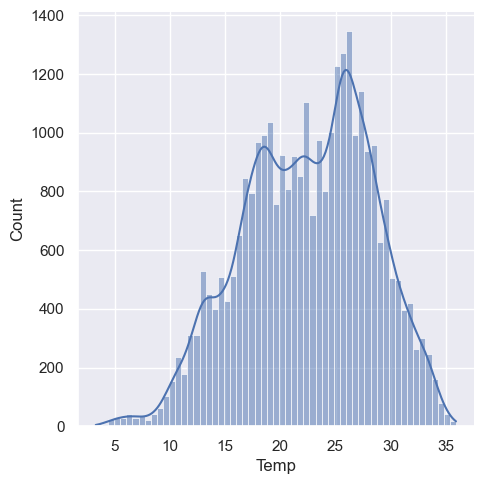}
		\caption{Temperature: histogram}\label{fig:histogram1}
  \end{minipage}
\begin{minipage}{0.3\textwidth}
		\centering
		\includegraphics[width=\textwidth]{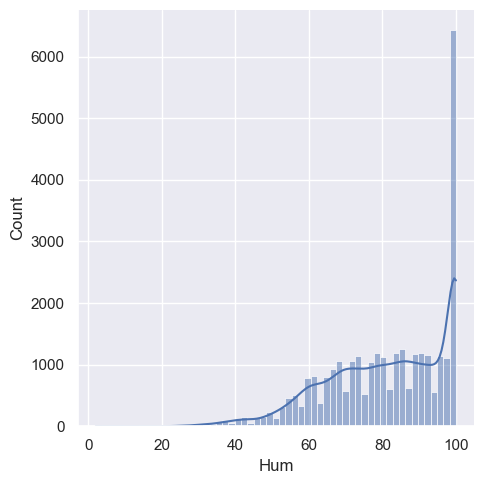}
		\caption{Humidity: histogram}\label{fig:histogram2}
  \end{minipage}
\begin{minipage}{0.3\textwidth}
		\centering
		\includegraphics[width=\textwidth]{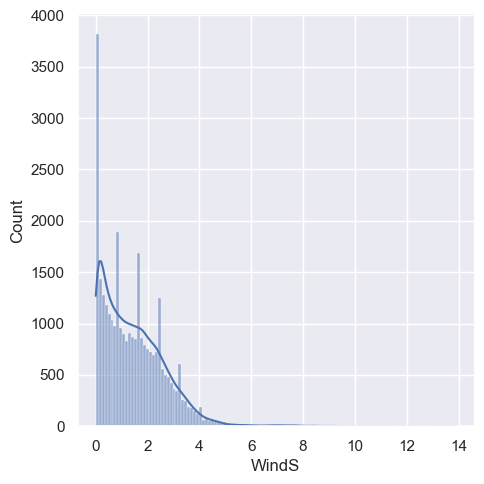}
		\caption{Wind speed: histogram}\label{fig:histogram3}
  \end{minipage}
\begin{minipage}{0.3\textwidth}
		\centering
		\includegraphics[width=\textwidth]{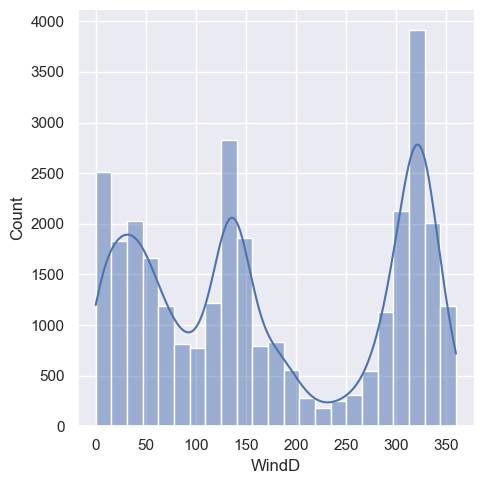}
		\caption{Wind direction: histogram}\label{fig:histogram4}
  \end{minipage}
\begin{minipage}{0.3\textwidth}
		\centering
		\includegraphics[width=\textwidth]{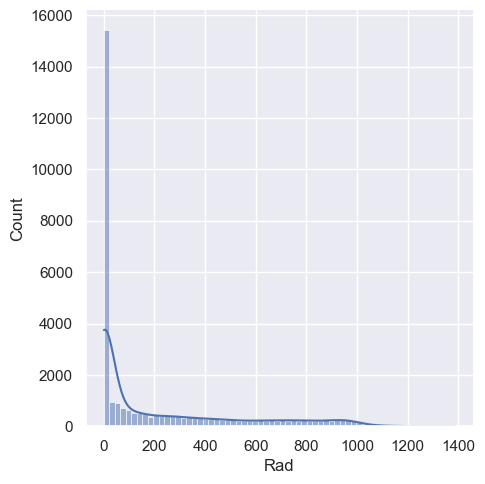}
		\caption{Radiation: histogram}\label{fig:histogram5}
\end{minipage}
\begin{minipage}{0.3\textwidth}
		\centering
		\includegraphics[width=\textwidth]{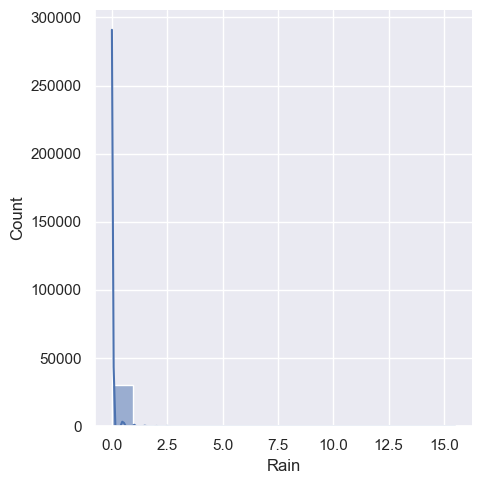}
		\caption{Rainfall: histogram}\label{fig:histogram6}
\end{minipage}
\begin{minipage}{0.3\textwidth}
		\centering
		\includegraphics[width=\textwidth]{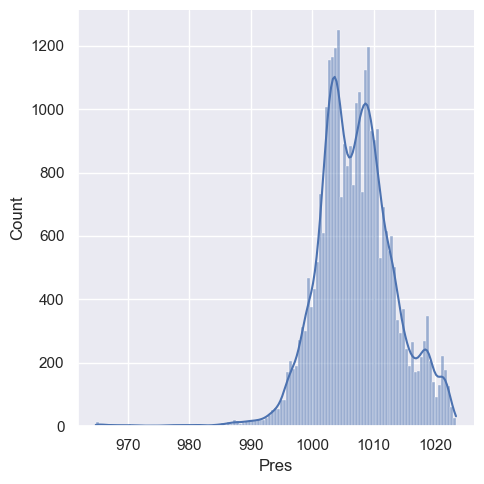}
		\caption{Pressure: histogram}\label{fig:histogram7}
\end{minipage}
\end{figure}

To provide a glimpse into the dataset, Table \ref{datasample} displays a sample of the recorded data, including the date and time of observation, as well as the corresponding values for each of the seven weather parameters.  Figures \ref{fig:histogram1},\ref{fig:histogram2},\ref{fig:histogram3},\ref{fig:histogram4},\ref{fig:histogram5},\ref{fig:histogram6},\ref{fig:histogram7} give the distribution of each weather variable in view of histogram in 2022.

\newpage

\begin{table}
	\centering
	\caption{A data sample in the dataset}
	\begin{tabular}{c c  c  c  c c }
		\toprule
		Time & Temperature  & Humidity & Wind speed & Wind direction & Radiation \\ 
		& [\textdegree{}C] & [$\%$] & [m/s] & [$\circ$] & [$W/m^2$]  \\ 
		\midrule
		21/10/15 0:00 & 17.70 & 94.0 & 0.000 & 0.0 & 0.00 \\ 
		21/10/15 0:10 & 17.60 & 94.0 & 0.200 & 323.0 & 0.00 \\ 
		21/10/15 0:20 & 17.70 & 94.0 & 0.100 & 337.0 & 0.00  \\ 
		21/10/15 0:30 & 17.70 & 93.0 & 0.000 & 317.0 & 0.00  \\ 
		21/10/15 0:40 & 17.90 & 91.0 & 0.500 & 325.0 & 0.00  \\ 
		21/10/15 0:50 & 17.70 & 91.0 & 0.100 & 320.0 & 0.00  \\ 
		21/10/15 1:00 & 17.80 & 92.0 & 0.100 & 302.0 & 0.00  \\ 
		21/10/15 1:10 & 17.80 & 92.0 & 0.000 & 0.0 &   0.00  \\ 
		21/10/15 1:20 & 18.30 & 89.0 & 0.500 & 327.0 & 0.00  \\ 
		21/10/15 1:30 & 17.90 & 89.0 & 0.300 & 11.0 & 0.00  \\ 
		21/10/15 1:40 & 18.10 & 89.0 & 0.200 & 344.0 & 0.00 \\ 
		21/10/15 1:50 & 18.20 & 89.0 & 0.500 & 306.0 & 0.00  \\ 
		21/10/15 2:00 & 18.10 & 88.0 & 0.200 & 300.0 & 0.00  \\ 
		21/10/15 2:10 & 18.10 & 88.0 & 0.300 & 299.0 & 0.00  \\ 
		\bottomrule \\
		 & Rainfall & Pressure \\ 
		& [mm/10min] & [hPa] \\ 
		\midrule
		 & 0.0 & 1011.7 \\ 
		 & 0.0 & 1011.7 \\ 
		 & 0.0 & 1011.6 \\ 
		& 0.0 & 1011.7 \\ 
		 & 0.0 & 1011.7 \\ 
		& 0.0 & 1011.6 \\ 
		& 0.0 & 1011.6 \\ 
		& 0.0 & 1011.6 \\ 
		& 0.0 & 1011.5 \\ 
		& 0.0 & 1011.4 \\ 
		& 0.0 & 1011.4 \\ 
		& 0.0 & 1011.2 \\ 
		& 0.0 & 1011.2 \\ 
		& 0.0 & 1011.1 \\ 
		\bottomrule		
	\end{tabular}
	\label{datasample}
\end{table}

Our dataset does not contain any missing values, which means that we can directly utilize the data for constructing our prediction model without needing to fill in any gaps. All seven parameters that we have collected are essential for our prediction model, and we will utilize them accordingly. However, as the values of each parameter may have significant differences, normalization is required to make the data easier to handle. We will discuss the normalization process further in the data preprocessing section.

Furthermore, to effectively utilize our dataset within a deep learning model, it is essential to partition the data into training and test sets. We will provide an extensive discussion of this process when we apply our prediction model, ensuring a comprehensive understanding of our model's performance evaluation.

\begin{figure}[H]
	\begin{center}
		\includegraphics[width=0.9\linewidth]{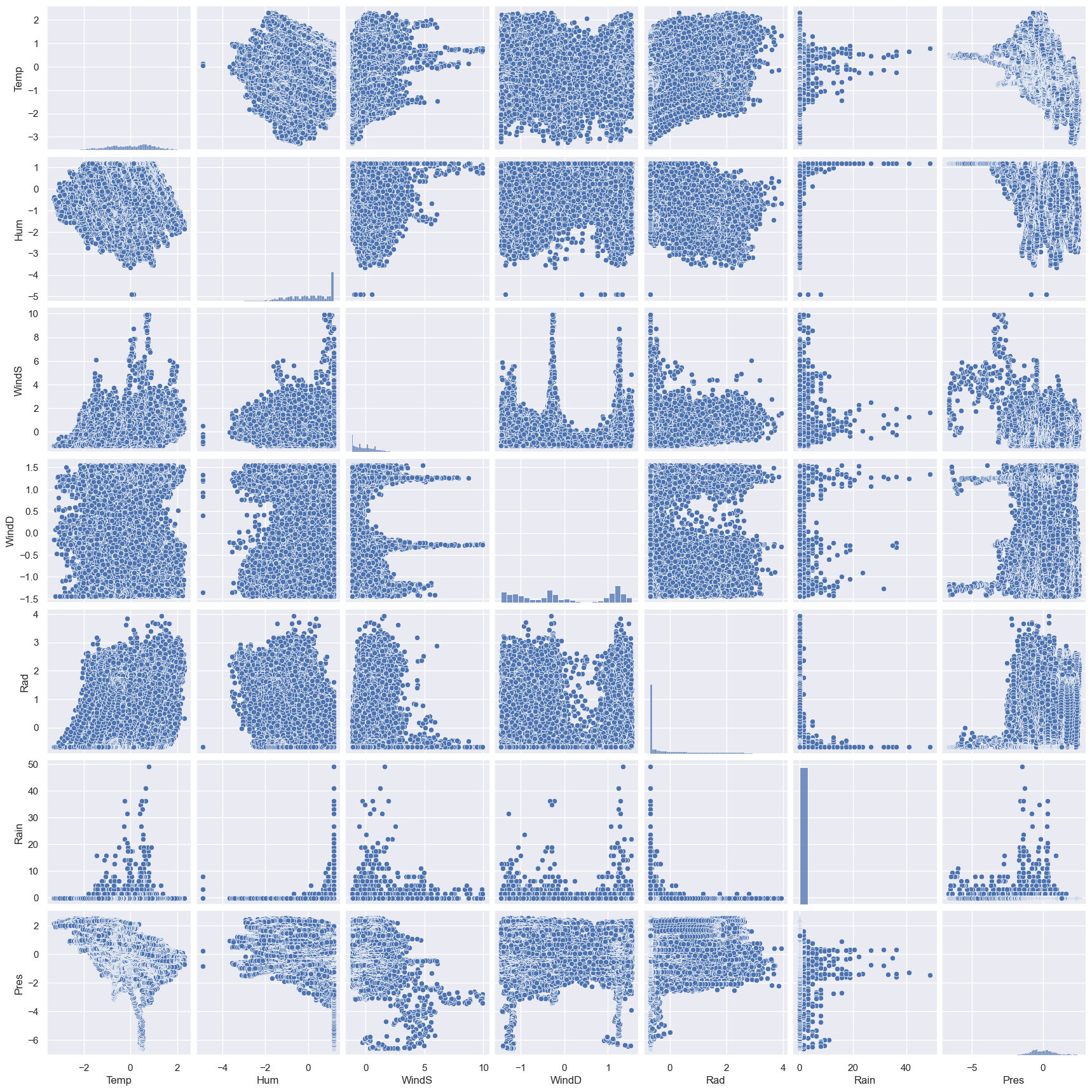}
		\caption{Correlation of weather parameters}
		\label{correlation}
	\end{center}
\end{figure}

This subsection explores the intricate relationships between various weather parameters and their correlations. Figure \ref{correlation} visually presents the correlations among these weather parameters.  Notably, the figure reveals that the relationships between the parameters are complex, lacking any discernible linear correlation between any pair of weather parameters. This observation underscores a critical point: traditional linear approaches, such as linear regression, may not be well-suited for modeling this dataset, given the intricate nonlinear interactions inherent in the data.

In response to this insight, our aim is to investigate a more sophisticated approach capable of handling the complex nonlinearities present in the dataset. To achieve this, we propose the application of deep neural networks to address the problem. We introduce the specific networks utilized and their associated algorithms in the remaining of this section.

\subsection{Multilayer perceptron: A brief overview}

Our primary approach for weather forecasting centers on the utilization of the Multilayer Perceptron (MLP), which is a type of deep neural network comprising linear networks and nonlinear activation functions.
Before delving into the specifics of our prediction model, let us give a preliminary introduction to MLP models.

An MLP model is composed of multiple layers with fully connected structure, usually an input layer for receiving data information, an output layer for producing the output of the network, and several hidden layers, layers between the input layer and output layers.
All layers consist of nodes and are connected through nodes from the previous and the next layers.
The complexity of the network is depending on the number of nodes and the number of layers.
Each node in an MLP receives input from other nodes in the preceding layer, performs a computation on that input using a linear map, and passes the result onto the next layer.
In order to introduce nonlinearity into the MLP and enable it to learn complex relationships in the data, activation functions are applied to the output of each node.
The MLP model is good at capturing intricate nonlinear relationships within data, making it an ideal choice for weather forecasting problems that exhibit complex nonlinearity. Also, the MLP model offers the added benefit of being able to predict all seven variables in our dataset simultaneously.
In following we give a brief introduction to MLP model with mathematical expression, more detailed explanation can be founded in \cite{zhang2021dive} and \cite{Roberts_2022}.

In MLP we usually use labelled data $(x_i,y_i)_{i=1}^n$ where $(x_i)_{i=1}^n$ are inputs, also called feature in deep learning, and $(y_i)_{i=1}^n$ be corresponding labels where each $y_i$ is a $d_{out}$-dimensional vector.
Each input $x_i$ a $d$ dimensional vector and the input layer with $d$ neurons consists of all components in $(x_i)_{i=1}^n$.
Suppose now we have $d'$ neurons in a layer of MLP. To obtain the $n_{out}$ neurons in a layer of MLP, we use a fully connected structure.
For each $d$-dimentional input $v$, the value $o_i$ at the $i$th neuron is computed by two steps: first, we calculate the preactivation $u_i$ by a linear combination of of one input $v = (v_j)_{j=1}^{d}$, weighted by $W =(w_{ij})\in \mathbb{R}^{d'\times d}$ and added to a bias term $b = (b_i)_{i=1}^{d'} \in \mathbb{R}^{d'}$; second, we apply a nonlinear activation function $\sigma(\cdot)$ to $u_i$ to obtain $o_i$.
The detailed expression is given by   
\begin{align*}
	&u_i = \sum_{j=1}^{d}w_{ij}v_j + b_i
	\\
	&o_i = \sigma(u_i) \,\,\,\, \operatorname{for} \, i = 1,2,...,d',
\end{align*}
then the output of this input $v$ is a $d_{out}$-dimensional vector $o$.
With this relation, given one input $x_k$, we can easliy define the MLP model with $M$ layers, with each layer $m$ composed of $d_m$ neurons, as 
\begin{align}
	&u_{ik}^{(1)} = \sum_{j=1}^{d}w_{ij}^{(1)}x_{jk} + b_i^{(1)} 
	\notag\\
	&o_{ik}^{(1)} = \sigma_1\left(u_{ik}^{(1)}\right) \,\,\,\, \operatorname{for} \, i = 1,2,...,d_1,
	\notag\\
	&u_{ik}^{(2)} = \sum_{j=1}^{d_1}w_{ij}^{(2)}o_{jk}^{(1)} + b_i^{(2)} 
	\notag\\
	&o_{ik}^{(2)} = \sigma_2\left(u_{ik}^{(2)}\right) \,\,\,\, \operatorname{for} \, i = 1,2,...,d_2,
	\notag\\
	&\, ... \\
	\notag
	&u_{ik}^{(M-1)} = \sum_{j=1}^{d_{M-2}}w_{ij}^{(M-1)}o_{jk}^{(M-2)} + b_i^{(M-1)} 
	\notag\\
	&o_{ik}^{(M-1)} = \sigma_m\left(u_{ik}^{(M-1)}\right) \,\,\,\, \operatorname{for} \, i = 1,2,...,d_{M-1}
	\notag\\
	&o_{ik}^{(M)} = \sum_{j=1}^{d_{M-1}}w_{ij}^{(M)}o_{jk}^{(M-1)} + b_i^{(M)} \,\,\,\, \operatorname{for} \, i = 1,2,...,d_{M}
	\label{MLP}
\end{align}
where $d_M = d_{out}$, $(o_{ik}^{(m)})_{i=1}^{d_m}$ are neurons in layer $m$, $\sigma_i$ are properly chosen activation functions, and $w_{ij}^{(m)}$ and $b_i^{(m)}$ are weights and biases in layer $m$. The most commonly used activation functions in neural networks are the ReLU and sigmoid functions, which are defined by the functions
\begin{align}
	&\sigma_{ReLU}(x) = \operatorname{max}\{x,0\}
	\label{relu}\\
	&\sigma_{sigmoid}(x) = \frac{1}{1+e^{-x}}.
	\label{sigmoid}
\end{align} 
\begin{figure}[H]
\begin{minipage}{0.45\textwidth}
		\includegraphics[width=0.95\linewidth]{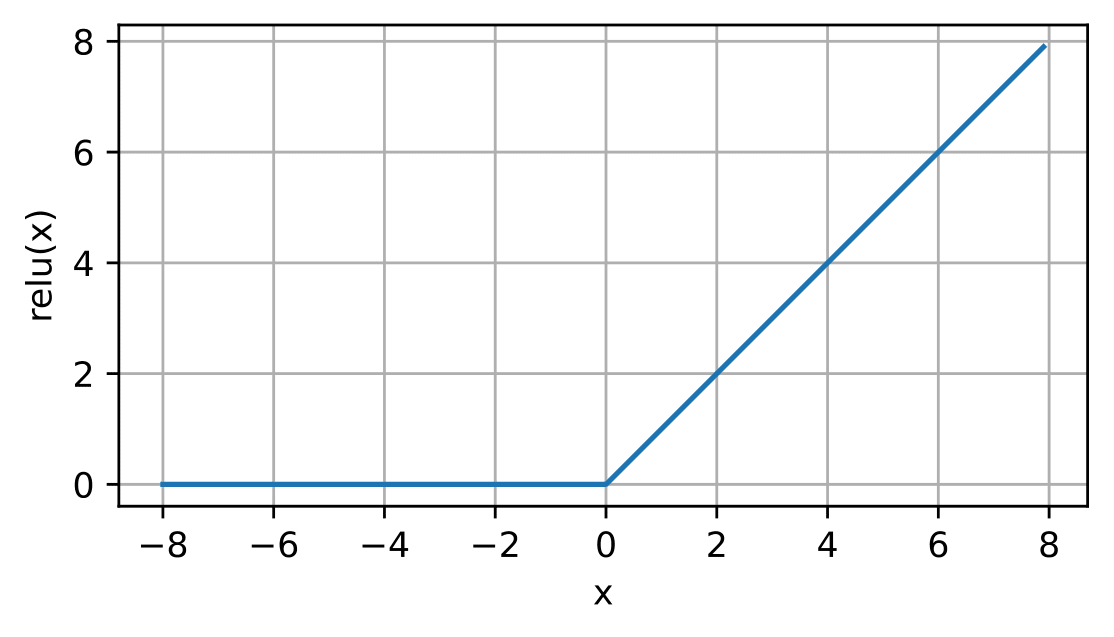}
		\caption{ReLU function}
		\label{relu_activation}
\end{minipage}
\hfill
\begin{minipage}{0.45\textwidth}
		\includegraphics[width=0.95\linewidth]{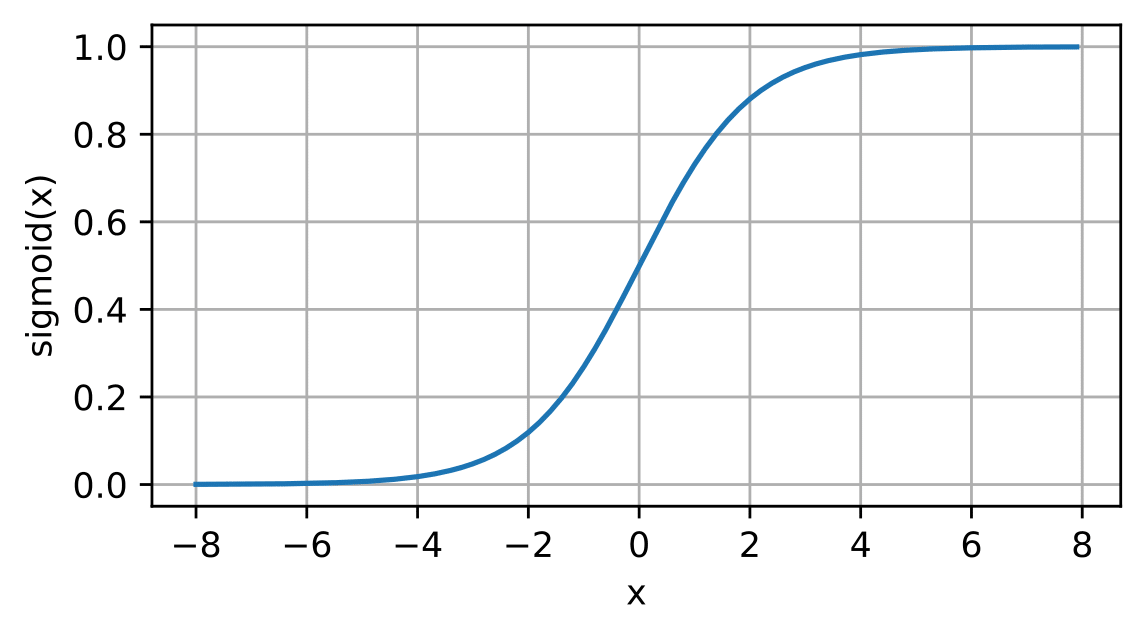}
		\caption{sigmoid function}
		\label{sig_activation}
  \end{minipage}
\end{figure}
Figures \ref{relu_activation} and \ref{sig_activation} show the nonlinearity of these two functions. 
In our case, layer $M$ refers to the output layer of the MLP, and the neurons $o^{(M)}$ provide the final output of the model. Observe that the output layer does not have an activation operation. 
Since each layer in the MLP model utilizes the full information of neurons from the previous layer, the structure is referred to as fully connected. 

\begin{figure}[H]
	\includegraphics[width=0.95\linewidth]{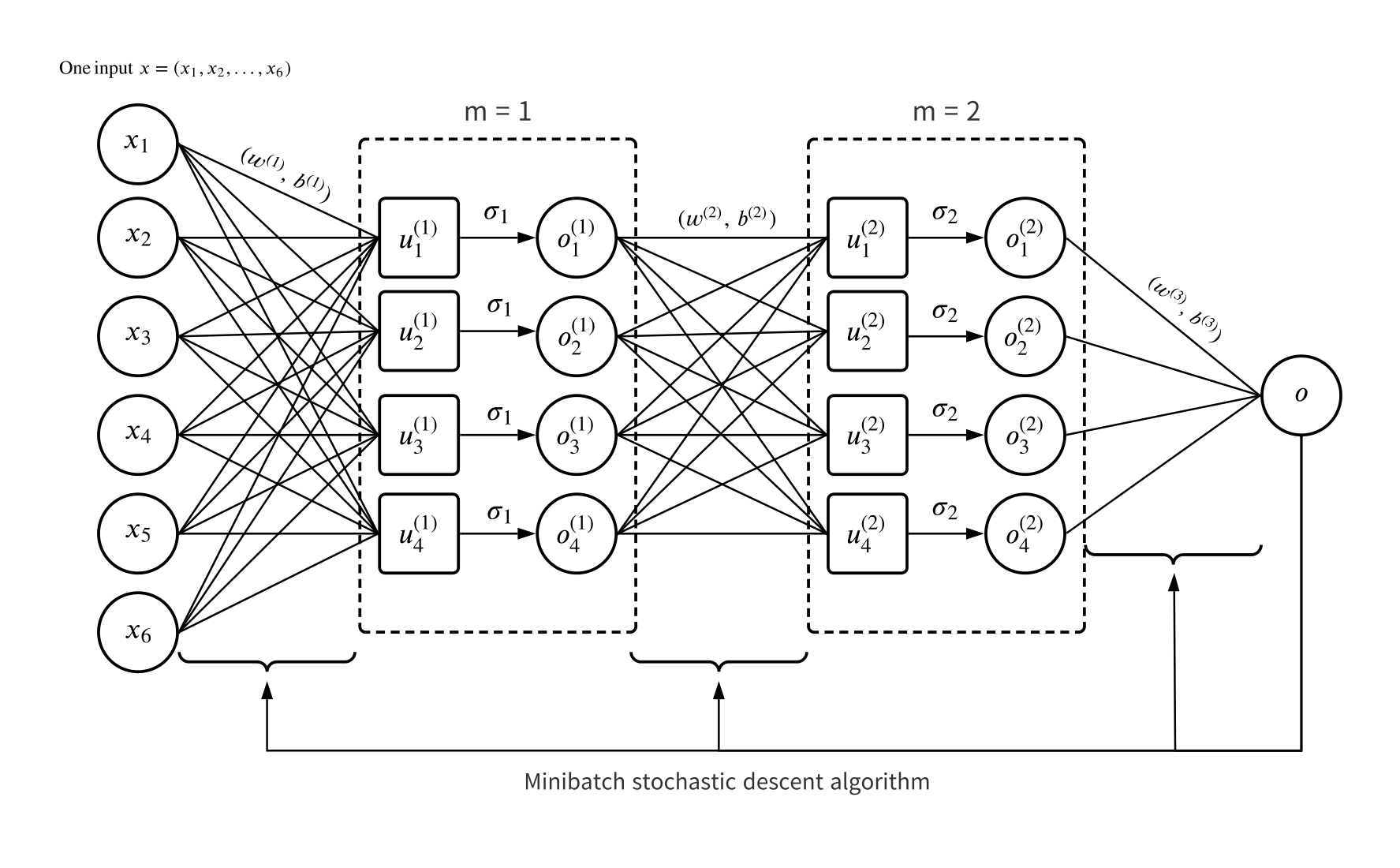}
	\caption{An example of MLP network consisting of 3 layers: input layer, output layer and two hidden layers.}
	\label{MLPfigure}
\end{figure}

In order to measure the accuracy of the network, we need to introduce a loss function. The most common loss function used in MLP model is the squared loss function, which is defined by 
\begin{align}
	L_n\left(y,o^{(M)}\right) &= \frac1n \sum_{k=1}^{n} l\left(y_k ,(o_{ik}^{(M)})_{i=1}^{d_M}\right) 
	\notag\\
	&= \frac1n \sum_{k=1}^{n} \sum_{i=1}^{d_{out}} \left(y_{ik}-o_{ik}^{(M)}\right)^2.
	\label{loss}
\end{align}
At the end of this overview, we introduce one method proposed \cite{li2014efficient} to update the parameters in \eqref{MLP} during the training procedure. The method is called minibatch stochastic descent algorithm, which is a improvement of usual stochastic descent algorithm.  
In each iteration, we first randomly choose a subset $\mathcal{Z}$ of all training data $(x_i,y_i)_{i=1}^n$ with predetermined cardinality $|\mathcal{Z}|$, we then let $\mathcal{Z}=(x_k,y_k)_{k=1}^{|\mathcal{Z}|}$ after a permutation of index.
Now the update of parameters is given by 
\begin{align}
	w_{ij}^{(m)} &\xleftarrow{} w_{ij}^{(m)} - \eta \partial_{w_{ij}^{(m)}} L_{|\mathcal{Z}|}\left(y,o^{(M)}\right)
	\notag\\
	&= w_{ij}^{(m)} - \frac{\eta}{|\mathcal{Z}|}\sum_{k=1}^{|\mathcal{Z}|} \partial_{w_{ij}^{(m)}} l\left(y_k,(o_{ik}^{(M)})_{i=1}^{d_M}\right)
	\\
	b_{i}^{(m)} &\xleftarrow{} b_{i}^{(m)} - \eta \partial_{b_{i}^{(m)}} L_{|\mathcal{Z}|}\left(y,o^{(M)}\right)
	\notag\\
	&= b_{i}^{(m)} - \frac{\eta}{|\mathcal{Z}|}\sum_{k=1}^{|\mathcal{Z}|} \partial_{b_{i}^{(m)}} l\left(y_k,(o_{ik}^{(M)})_{i=1}^{d_M}\right)
\end{align}
where $\eta$ is a small positive value, called learning rate. In this algorithm, $\eta$ and $|\mathcal{Z}|$ are chosen before applying algorithm, so they are hyperparameters. We will introduce  a method to choose efficient hyperparameter when we apply this algorithm.
To better visualize the whole picture of the MLP network, we provide an example in Figure \ref{MLPfigure}.

\subsection{Recurrent neural network and Long short term memory} 
\begin{figure}[H]
	\includegraphics[width=0.95\linewidth]{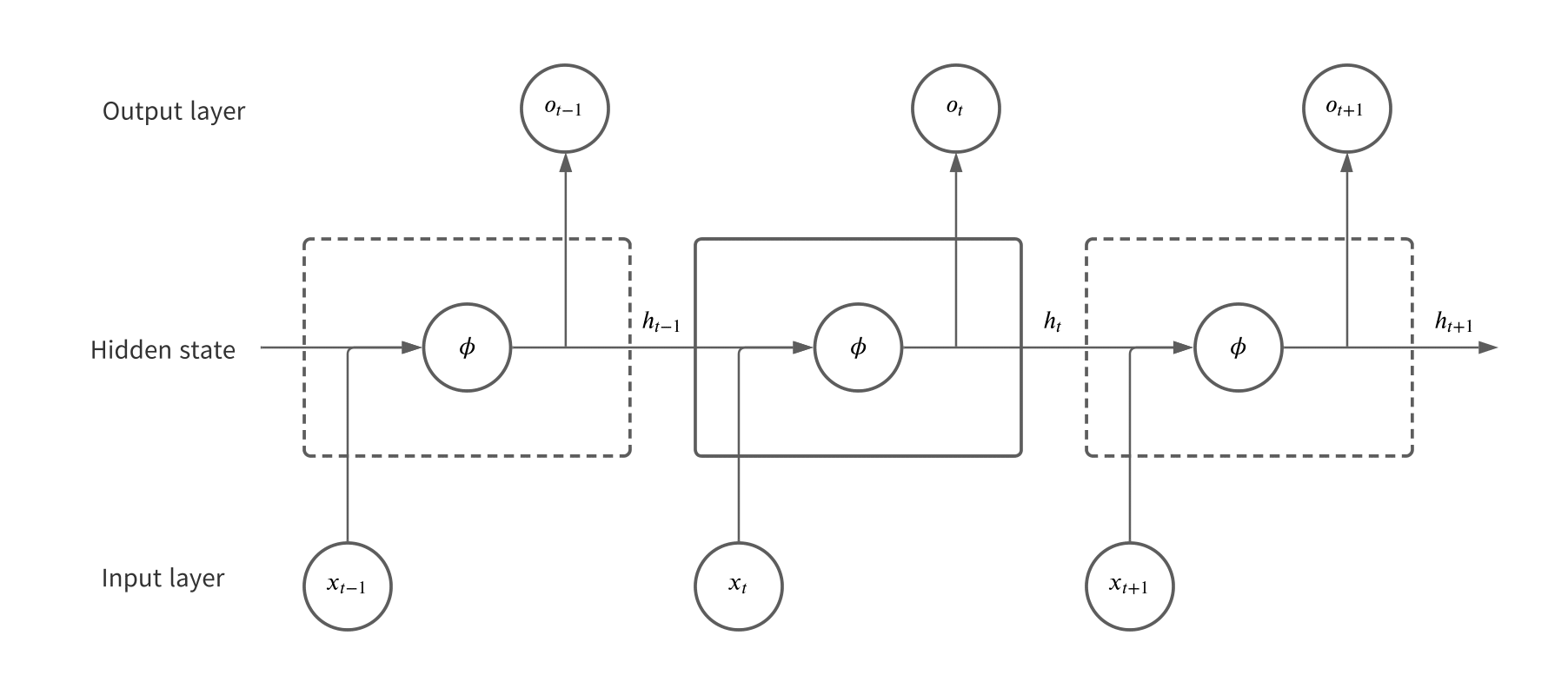}
	\caption{A RNN cell}
	\label{RNNcell}
\end{figure}

In this subsection, we provide a concise introduction to another widely used deep neural network model known as Recurrent Neural Network (RNN). RNN is specifically designed to handle time series tasks, and one of its specialized variants is Long Short-Term Memory (LSTM).
Typically, RNN models are fed with sequential data as input. Each RNN layer consists of multiple RNN cells, the number of which corresponds to the length of the input sequence. Within each RNN cell, the input value $x_t$ is combined with the hidden state $h_{t-1}$ computed from the previous time step. The operation within each RNN cell is depicted in Figure \ref{RNNcell}, and mathematically expressed as follows:
\begin{align}
	h_t &= \phi\left(w_1 x_t + w_2 h_{t-1} +b_h\right)
	\notag\\
	o_t &= w_oh_t + b_o
	\label{cell_eq}
\end{align} Here, $w_1$, $w_2$ are the weights, and $b_h$ represents the bias in the hidden state layer. Similarly, $w_o$ and $b_o$ denote the weight and bias in the output layer. The activation function $\phi$ introduces non-linearity into the model, allowing it to capture complex relationships within the data.

\begin{figure}[H]
	\includegraphics[width=0.75\linewidth]{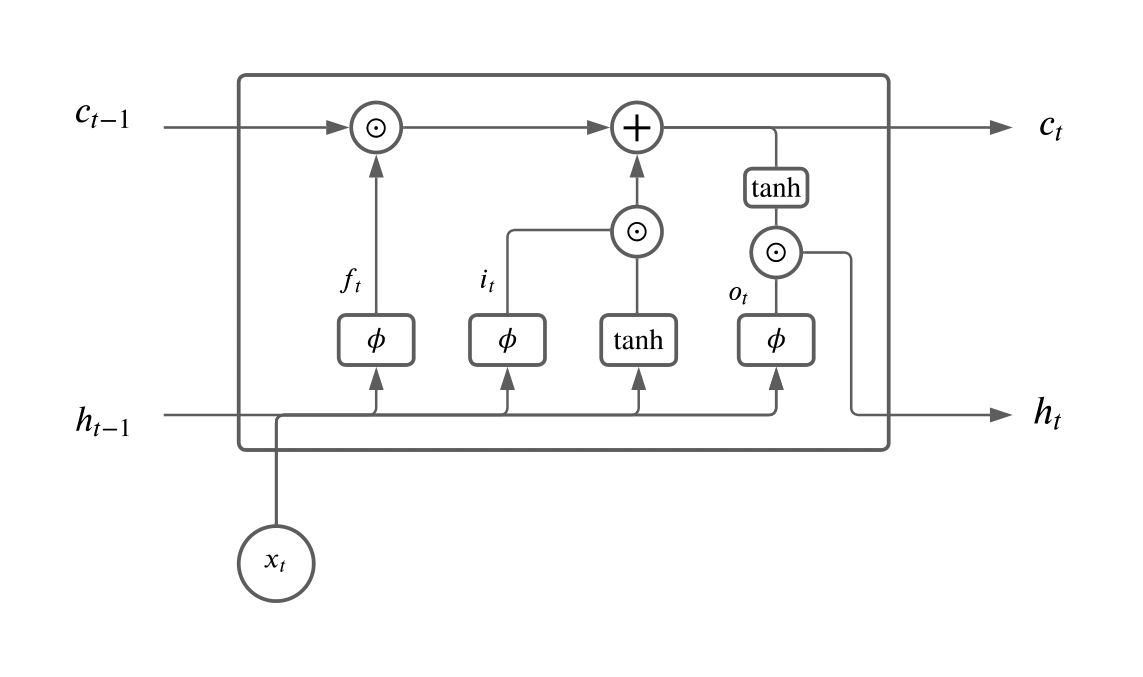}
	\caption{A LSTM cell}
	\label{LSTMcell}
\end{figure}

LSTM is a specialized variant of the RNN model, characterized by three gate operations: the forget gate, input gate, and output gate, as well as the presence of an additional state called the internal state $c_t$, which preserves current memory during computation.
The forget gate determines how much of the long-term memory should be retained and is computed as follows:
\begin{equation*}
	f_t = \phi\left(w_{f,1}x_t+w_{f,2}h_{t-1}+b_f\right)
\end{equation*} where $w_{f,1}$, $w_{f,2}$ are weight and $b_f$ is bias.
The input gate calculates the information to be added to the memory and is given by:
\begin{equation*}
	i_t = \phi\left(w_{i,1}x_t+w_{i,2}h_{t-1}+b_i\right).
\end{equation*} 
Similarly, the output gate is determined by:
\begin{equation*}
	o_t = \phi\left(w_{o,1}x_t+w_{o,2}h_{t-1}+b_o\right).
\end{equation*} 
The internal state $c_t$ is computed by the equation:
\begin{equation*}
	c_t = f_t \odot c_{t-1} + i_t \odot \tanh(h_{t-1})
\end{equation*} where operator $\odot$ denotes Hadamard (elementwise) product. The first term on the right-hand side represents the long-term memory, while the second term represents the new information added to the state.
Finally, the hidden state in the cell, denoted by $h_t$, is given by:
\begin{equation*}
	h_t = o_t \odot \tanh(c_t).
\end{equation*} 
We illustrate the computation flow of LSTM cell by Figure \ref{LSTMcell}.
The parameter update algorithm for RNN is a variation of the Backward Propagation algorithm, known as Backpropagation Through Time (BTT). 
For detailed insights into BTT and its implementation, Readers are recommend referring to the source, \cite{zhang2021dive}.

\subsection{Our method}

In this subsection, we present our approach to weather forecasting based on MLP model. 
We use a 3 layer MLP model, consisting of one input layer, one hidden layer and one output layer. 
The number of nodes in input layer and output layer are determined by our data, while the number of nodes in hidden layer which determines the architecture of our network is a hyperparameter that we adjust during our process.
For the activation function of our network, we use the ReLU function \eqref{relu}. 
As in \eqref{MLP}, we have weights and biases in the network.
We initialize the weight parameter to follow a Normal distribution with mean zero and variance $0.01$, the bias parameter to be zero vector. Then we update them by using Minibatch stochastic descent algorithm with squared loss function \eqref{loss}. Another hyperparameter learning rate is also determined during our process.
The reason to choose 3 layers is that adding more layers does not yield a significant improvement in the test accuracy for our dataset. In fact, more layers sometimes result in lower accuracy.

\begin{figure}[H]
	\includegraphics[width=0.95\linewidth]{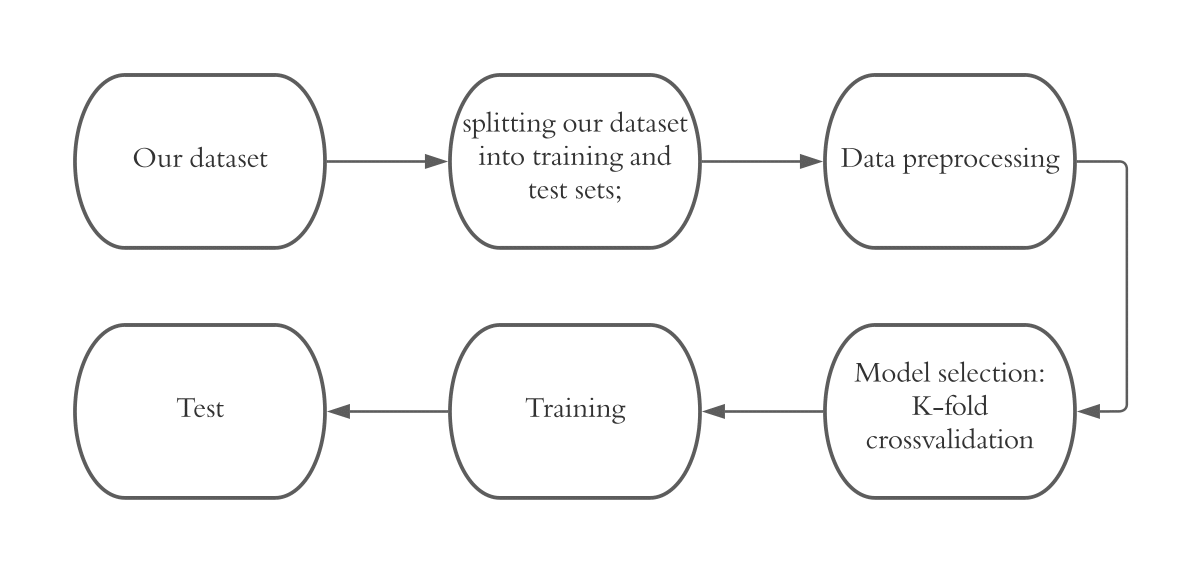}
	\caption{Flowchart of our method}
	\label{Flowfigure}
\end{figure}

After introducing the structure of our network, we give the flow of our approach. It consists of five main steps:
\begin{enumerate}
	\item Splitting our dataset into training and test sets;
	\item Preprocessing our data to make it more suitable for network use;
	\item Selecting an efficient pair of hyperparameters: learning rate and number of hidden neurons, by using K-fold cross-validation;
	\item Training our network on the training set;
	\item Evaluating prediction performance on the test set using a loss function.
\end{enumerate}
Figure \ref{Flowfigure} shows the flowchart of our approach.

Now we detail each step in our method.
\subsubsection{Training and test set}
We split our data set into training set and test set. The training set used in this study consists of 30816 observations collected in 2022, while the test set comprise 6769 observations collected in 2021. This approach was adopted to ensure that the model could better capture the periodic structure of the weather data, which varies from year to year.

\subsubsection{Data preprocessing}
Recall that an MLP network uses supervised learning, which means the data consists of both features and labels. Therefore we have to define features and labels for our data set. 

Our method is to take 3 consecutive 7-dimensional observations and flatten them into a single 21-dimensional vector. This vector is considered one feature. The observation immediately following these 3 observations is considered one label. We repeat this process for all observations in the training set. This results in 30813 21-dimensional features and 30813 7-dimensional labels for training. We can represent these as $(x_j)_{j=1}^{30813}$ and $(y_j)_{j=1}^{30813}$ respectively, so the training set is $(x_j,y_j)_{j=1}^{30813}$. By doing the same thing for the test set, we can obtain a labeled test set.

To better fit the data during the training process, we propose a method to first normalize the data before put it into our network. The method incorporated in our preprocessing is called min-max scaler, see \cite{endalie2022deep}. Given a set of observations $(x_j)_{j=1}^n$ with each $x_j=(x_{j1},x_{j2},...,x_{jd})$, then min-max scaler of $x_j$ is defined by
\begin{align}
	\Tilde{x}_{jk} = \frac{x_{jk}-\operatorname{min}\{x_{ik}; i=1,2,...,n \}}{\operatorname{max}\{x_{ik}; i=1,2,...,n \}-\operatorname{min}\{x_{ik}; i=1,2,...,n \}} 
	\notag
\end{align}for $k=1,...,d$.
We apply this min-max scaling normalization to both our training and test sets to obtain data that is suitable for our network. We should mention here that we normalize both the training and test sets using the maximum and minimum values in the training set.

\subsubsection{Model selection}

\begin{figure}[H]
	\includegraphics[width=0.75\linewidth]{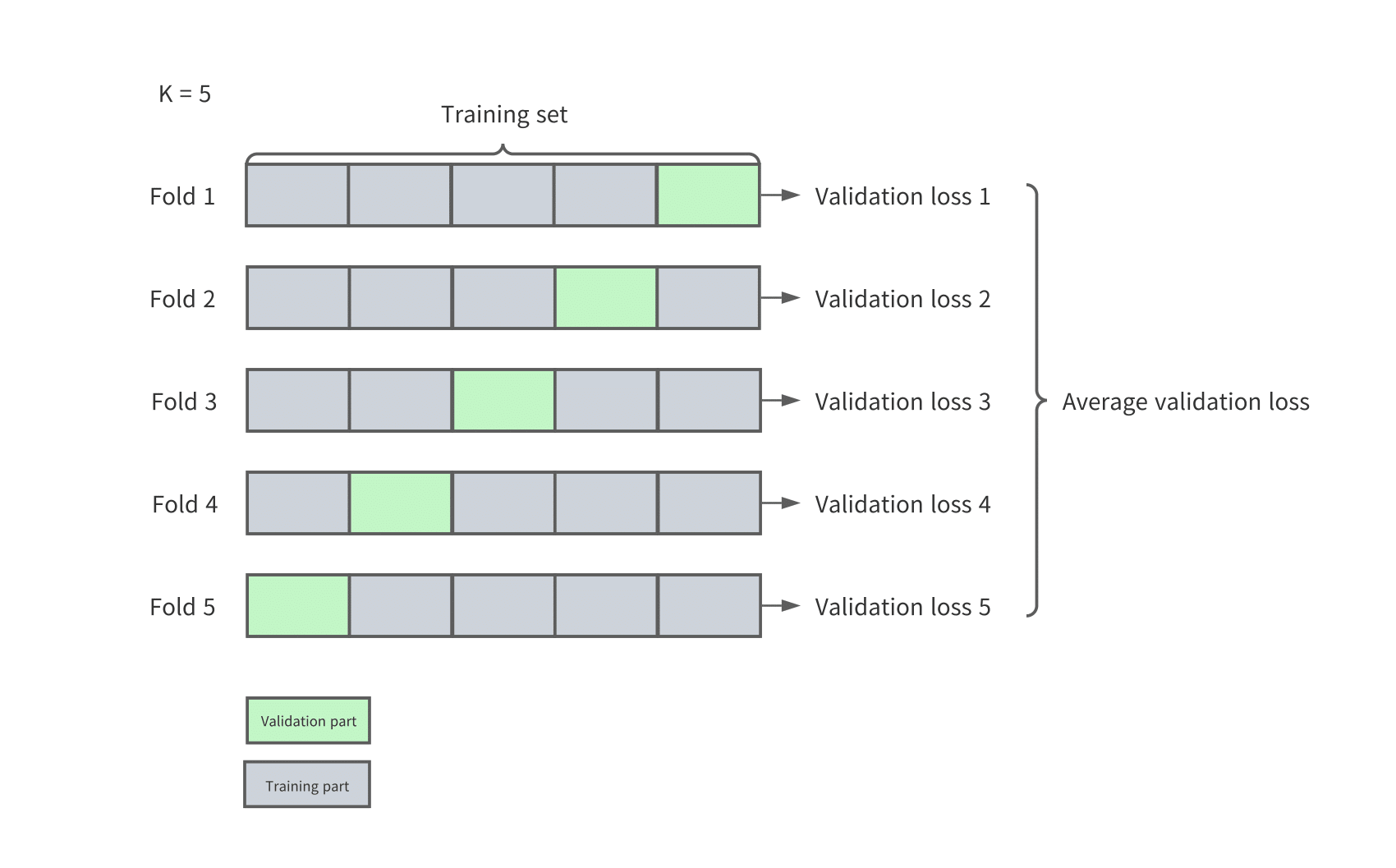}
	\caption{K-fold crossvalidation for $K = 5$.}
	\label{Crossfigure}
\end{figure}

As mentioned above, the learning rate and the number of nodes in the hidden layer are two important hyper factors in our model. The learning rate affects how quickly our algorithm learns, while the number of nodes in the hidden layer affects the structure of our network and its accuracy. So we want to determine a optimal choice of the pair $(\eta,n_{hidden})$, the learning rate $\eta$ and the number of hidden nodes $n_{hidden}$, to minimize the loss on our training set. This means choosing the right learning rate and number of hidden nodes to achieve the best results.
To find the best pair of learning rate and number of hidden nodes, we use a technique called K-fold cross-validation. This technique only runs on the training set. We start by creating a set of pairs, where each pair represents a different network structure. Then, we split the training set into K parts, with $K-1$ parts for training and the remaining part for validation. We repeat this process K times to create K different folds of data. For each fold, we train the network using one specific pair on the training part and test its performance on the validation part to calculate the validation loss. We then average the validation loss across all K folds to get an average validation loss for that pair. Finally, we repeat this process for all pairs in our set and choose the pair with the smallest average validation loss as the optimal choice for our network. See Figure \ref{Crossfigure} for an example.

In our case, we apply a 5-fold crossvalidation for set of pairs $S$, where
\begin{align*}
	S =
	\Big\{(\eta,n_{hidden})\colon &
	n_{hidden}\in \{32, 40, 48,\dots, 1016, 1024\}, \\
    &  \eta \in \{0.01,0.05,...,0.75, 0.8\}\Big\}.
\end{align*} 
The 5-fold crossvalidation select the pair $(0.7,344)$ as the optimal choice for our network. 

\subsubsection{Training and evaluation}

After completing the necessary steps in data preprocessing, including obtaining a well-labelled training set and test set, we applied a 5-fold cross-validation to determine the architecture of our network. 
We found that a Multilayer Perceptron (MLP) network with 3 layers and 344 hidden nodes would be the most appropriate for our purposes.
Now we proceed to train our network on the training set using the minibatch stochastic descent algorithm with a squared loss function \eqref{loss} and a learning rate of 0.7. The algorithm iterates several times to minimize the training loss until it converges to a stable state. Once the algorithm reaches this stable state, it stops iterating and provides the final trained model.
To evaluate the accuracy of our model, we use the test set as input data into the trained model and calculate the accuracy using the squared loss function \eqref{loss}. The resulting test accuracy provides a measure of the performance of the trained model. Additionally, we trained the RNN and LSTM models to provide a comparison with our model. The training strategy for both the RNN and LSTM models was the same as for our model.

\section{Results}

This section presents the outcomes of our network's performance on the labeled dataset introduced in the previous section. First, we initialize the network's parameters by assigning all weight parameters as normal random variables with a mean of 0 and a variance of 0.01, while bias parameters are set to zeros.

Furthermore, we present a comparative analysis of our model alongside two types of RNN models introduced in the preceding subsection: the Simple RNN model and the LSTM model. Both the Simple RNN and LSTM models employ the same training and test data as our model, with their respective hyperparameters selected through validation loss.

\begin{figure}
 \begin{minipage}{0.48\textwidth}
 \centering
		\includegraphics[width=\linewidth]{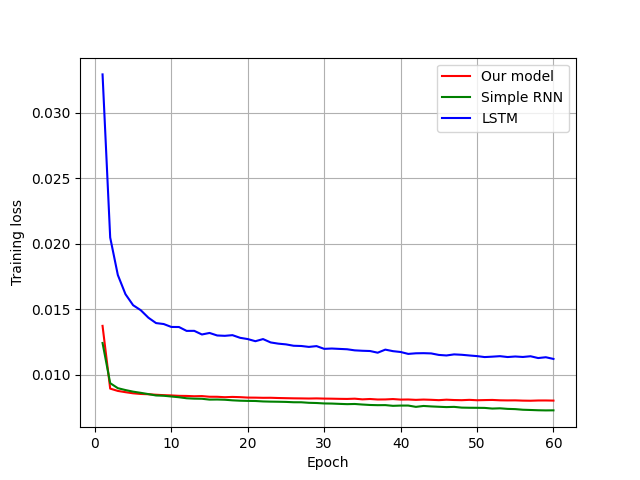}
		\caption{Training loss in 60 epochs.}
		\label{Trainingloss}\label{fig:loss and accuracy1}
\end{minipage}
\begin{minipage}{0.48\textwidth}
\centering
		\includegraphics[width=\linewidth]{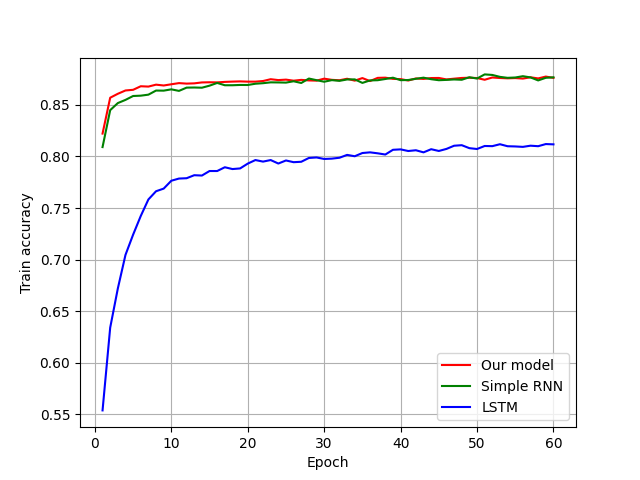}
		\caption{Training accuracy in 60 epochs.}
		\label{Trainingaccuracy}\label{fig:loss and accuracy2}
  \end{minipage}
  \hfill
\begin{minipage}{0.48\textwidth}
\centering
		\includegraphics[width=\linewidth]{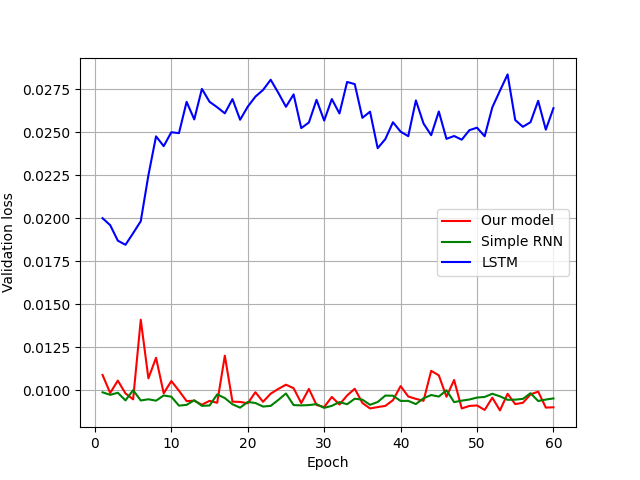}
		\caption{Validation loss in 60 epochs.}
		\label{Validationloss}\label{fig:loss and accuracy3}
  \end{minipage}
\begin{minipage}{0.48\textwidth}
\centering
		\includegraphics[width=\linewidth]{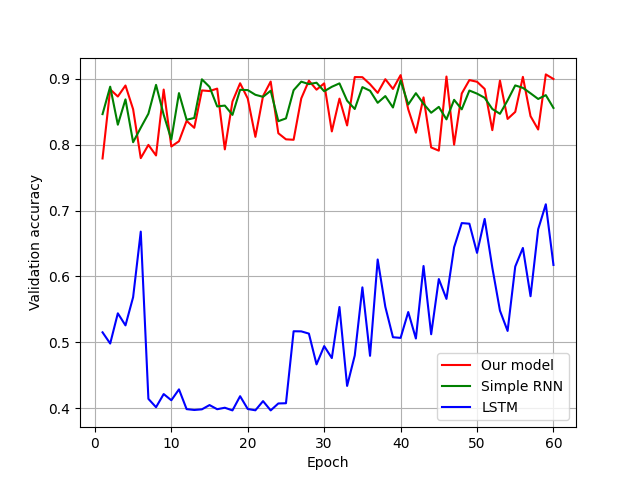}
		\caption{Validation accuracy in 60 epochs.}
		\label{Validationaccuracy}\label{fig:loss and accuracy4}
\end{minipage}
\end{figure}

The training process of our network adhered to the configurations detailed in Table \ref{tab:networkconfig}. The training performance, including training loss, training accuracy, validation loss, and validation accuracy, for our network, as well as the RNN and LSTM models, is visually represented in Figures \ref{fig:loss and accuracy1}-\ref{fig:loss and accuracy4}. It's worth noting that our network and RNN achieved favorable training loss on the training set, while the LSTM models exhibited lower accuracy in comparison.

\begin{figure}[H]
\centering
\begin{minipage}{0.33\textwidth}
		\includegraphics[width=\linewidth]{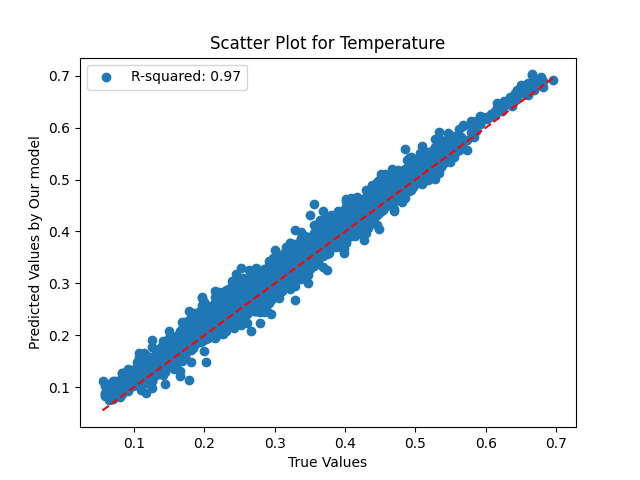}
		\caption{Temperature\\Our model}
		\label{TemMLP}
  \end{minipage}
 \begin{minipage}{0.33\textwidth}
		\includegraphics[width=\linewidth]{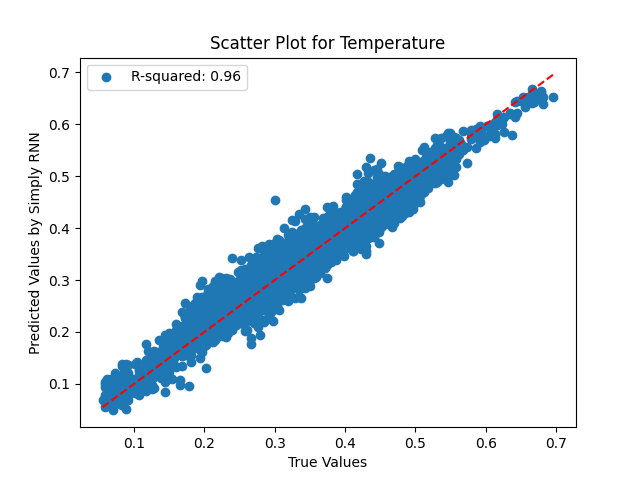}
		\caption{Temperature\\Simple RNN}
		\label{TemRNN}
	\end{minipage}
 \begin{minipage}{0.33\textwidth}
		\includegraphics[width=\linewidth]{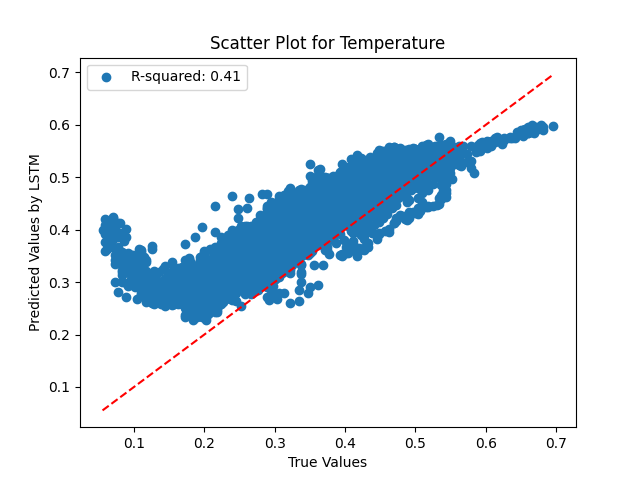}
		\caption{Temperature\\LSTM}
		\label{TemLSTM}
	\end{minipage}
 \hfill
 \begin{minipage}{0.33\textwidth}
		\includegraphics[width=\linewidth]{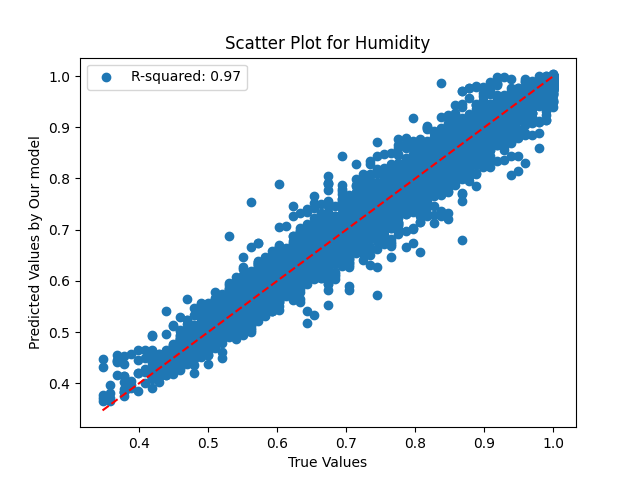}
		\caption{Humidity\\Our model}
		\label{HumidityMLP}
	\end{minipage}
 \begin{minipage}{0.33\textwidth}
		\includegraphics[width=\linewidth]{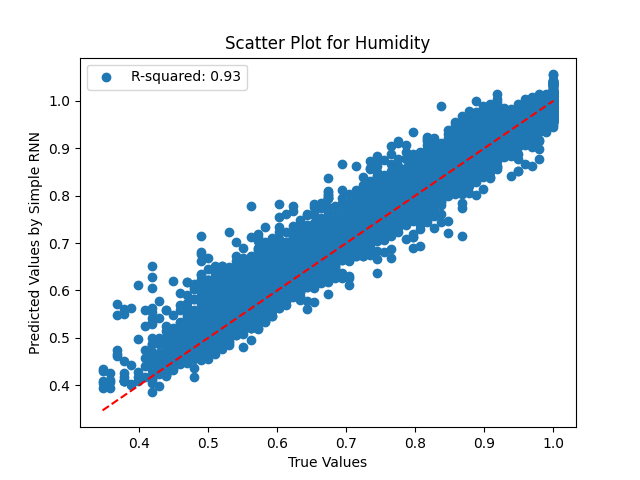}
		\caption{Humidity\\Simple RNN}
		\label{HumidityRNN}
	\end{minipage}
 \begin{minipage}{0.33\textwidth}
		\includegraphics[width=\linewidth]{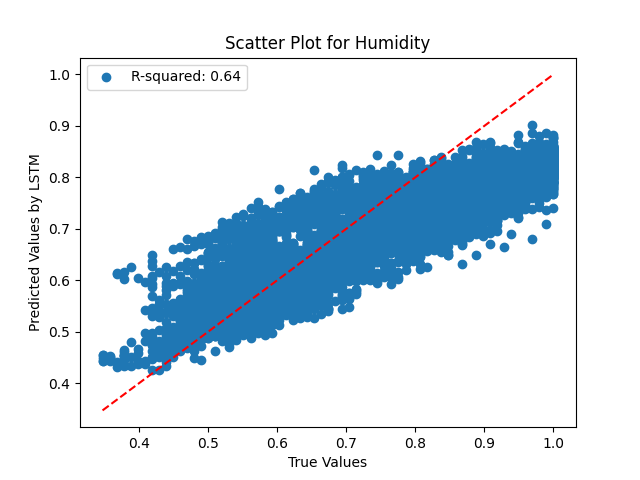}
		\caption{Humidity\\LSTM}
		\label{HumidityLSTM}
	\end{minipage}
 \hfill
 \begin{minipage}{0.33\textwidth}
		\includegraphics[width=\linewidth]{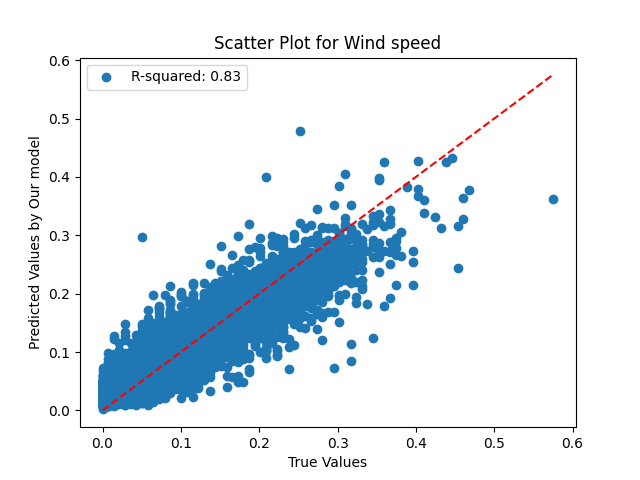}
		\caption{Wind speed\\Our model}
		\label{wsMLP}
	\end{minipage}
 \begin{minipage}{0.33\textwidth}
		\includegraphics[width=\linewidth]{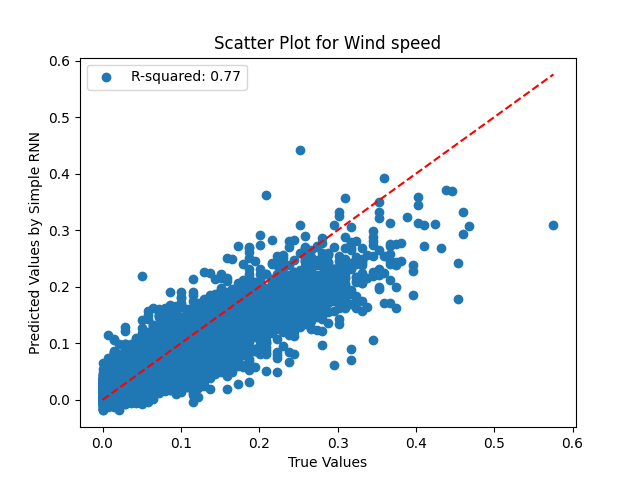}
		\caption{Wind speed\\Simple RNN}
		\label{wsRNN}
	\end{minipage}
 \begin{minipage}{0.33\textwidth}
		\includegraphics[width=\linewidth]{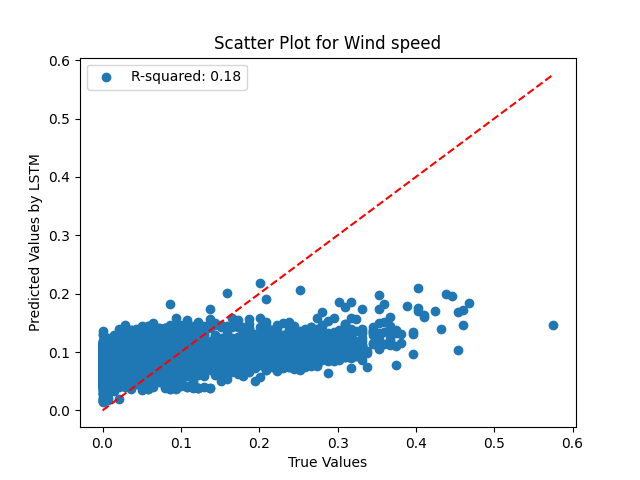}
		\caption{Wind speed\\LSTM}
		\label{wsLSTM}
	\end{minipage}
 \hfill
 \begin{minipage}{0.33\textwidth}
		\includegraphics[width=\linewidth]{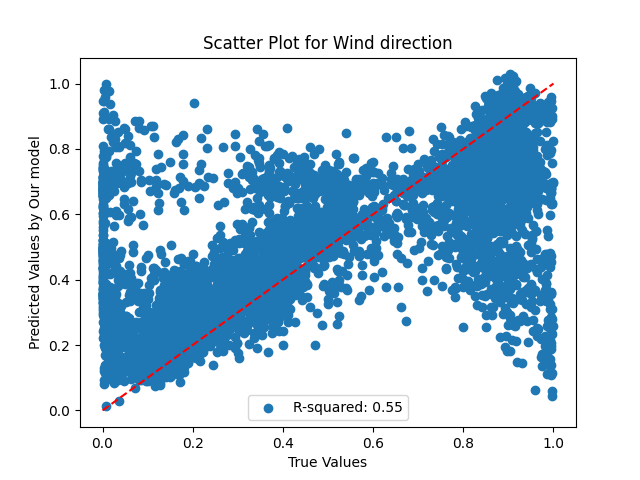}
		\caption{Wind direction\\Our model}
		\label{wdMLP}
	\end{minipage}
 \begin{minipage}{0.33\textwidth}
		\includegraphics[width=\linewidth]{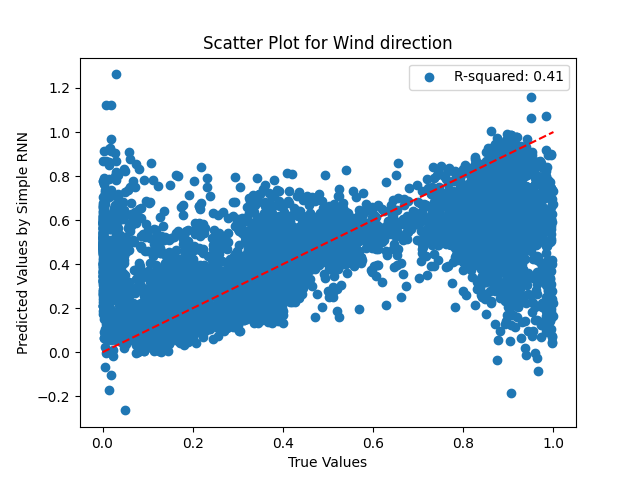}
		\caption{Wind direction\\Simple RNN}
		\label{wdRNN}
	\end{minipage}
 \begin{minipage}{0.33\textwidth}
		\includegraphics[width=\linewidth]{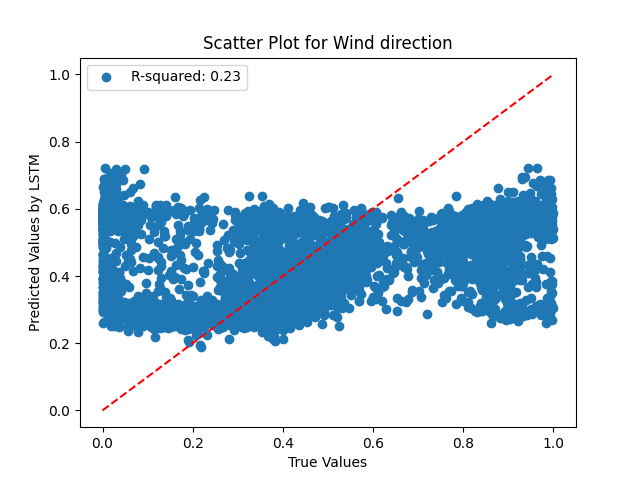}
		\caption{Wind direction\\LSTM}
		\label{wdLSTM}
	\end{minipage}
 \end{figure}

\begin{figure}[H]
     \centering
 \begin{minipage}{0.33\textwidth}
		\includegraphics[width=\linewidth]{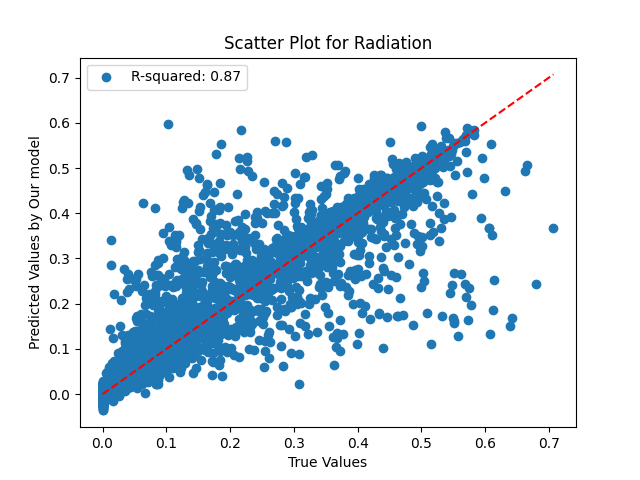}
		\caption{Radiation\\Our model}
		\label{RadiationMLP}
	\end{minipage}
 \begin{minipage}{0.33\textwidth}
		\includegraphics[width=\linewidth]{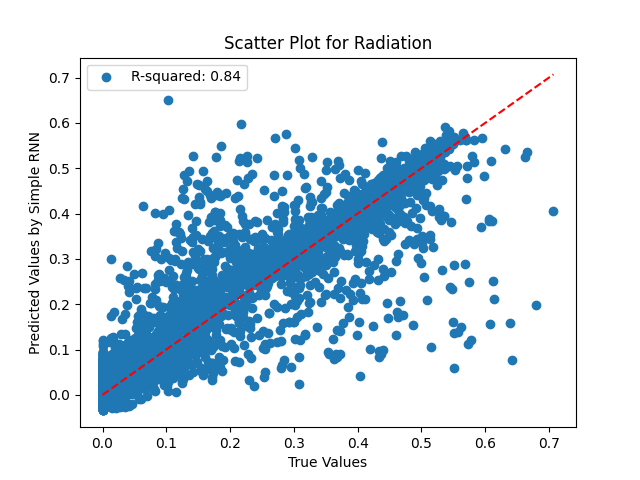}
		\caption{Radiation\\Simple RNN}
		\label{RadiationRNN}
	\end{minipage}
 \begin{minipage}{0.33\textwidth}
		\includegraphics[width=\linewidth]{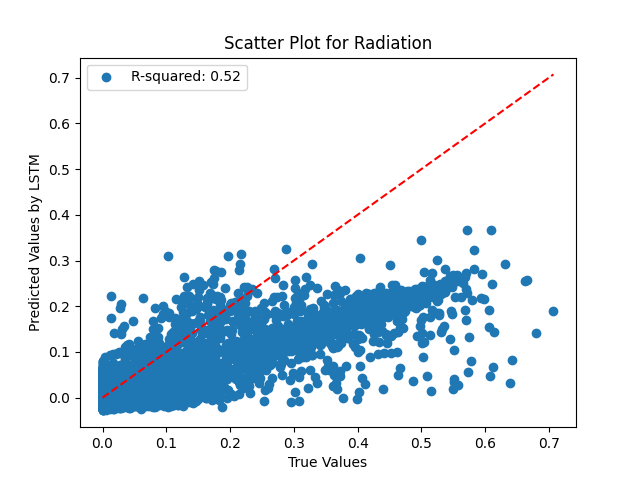}
		\caption{Radiation\\LSTM}
		\label{RadiationLSTM}
	\end{minipage}
 \hfill
 \begin{minipage}{0.33\textwidth}
		\includegraphics[width=\linewidth]{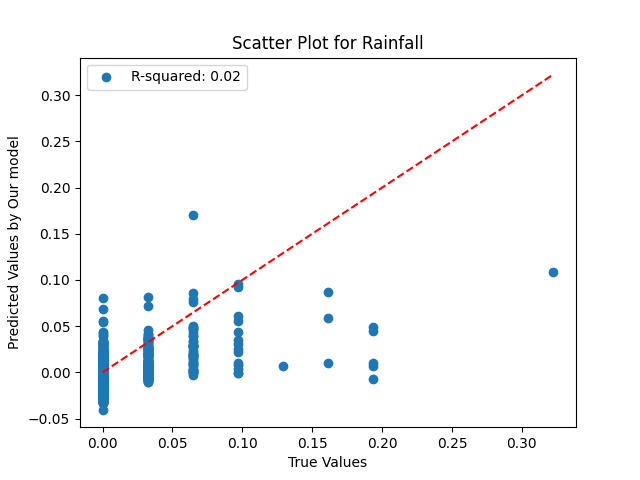}
		\caption{Rainfall\\Our model}
		\label{RainfallMLP}
	\end{minipage}
 \begin{minipage}{0.33\textwidth}
		\includegraphics[width=\linewidth]{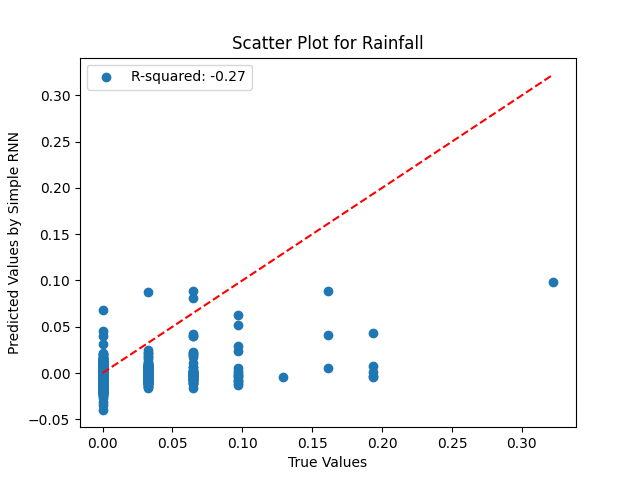}
		\caption{Rainfall\\Simple RNN}
		\label{RainfallRNN}
	\end{minipage}
 \begin{minipage}{0.33\textwidth}
		\includegraphics[width=\linewidth]{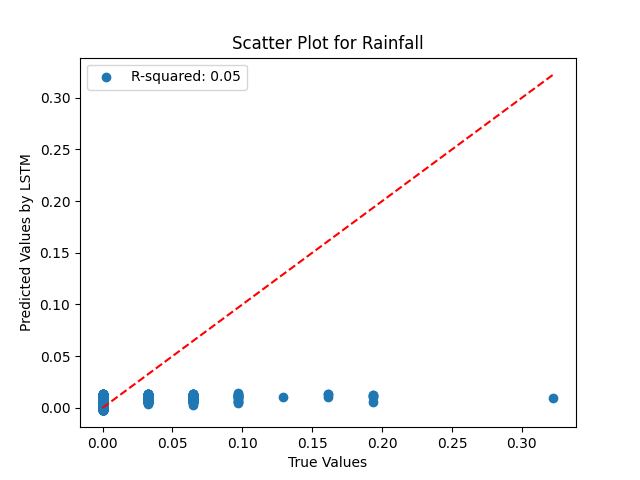}
		\caption{Rainfall\\LSTM}
		\label{RainfallLSTM}
	\end{minipage}
 \hfill
 \begin{minipage}{0.33\textwidth}
		\includegraphics[width=\linewidth]{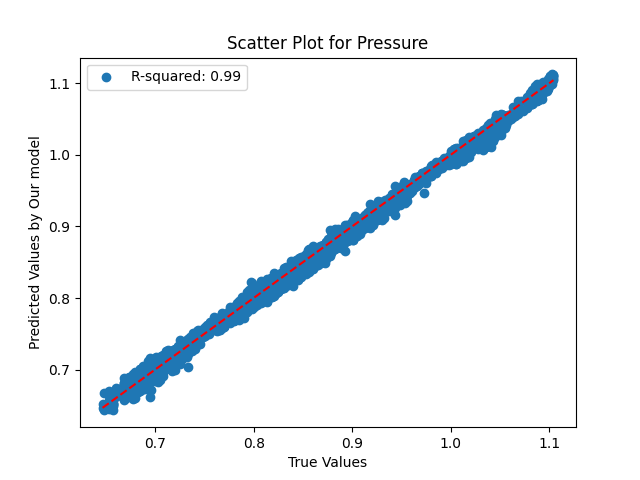}
		\caption{Pressure\\Our model}
		\label{PressureMLP}
	\end{minipage}
 \begin{minipage}{0.33\textwidth}
		\includegraphics[width=\linewidth]{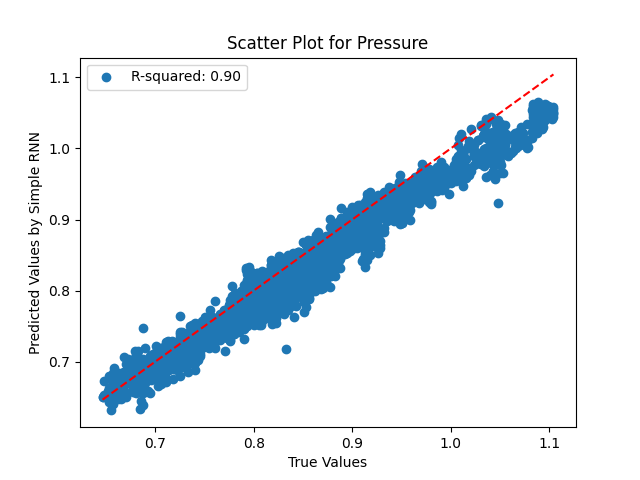}
		\caption{Pressure\\Simple RNN}
		\label{PressureRNN}
	\end{minipage}
 \begin{minipage}{0.33\textwidth}
		\includegraphics[width=\linewidth]{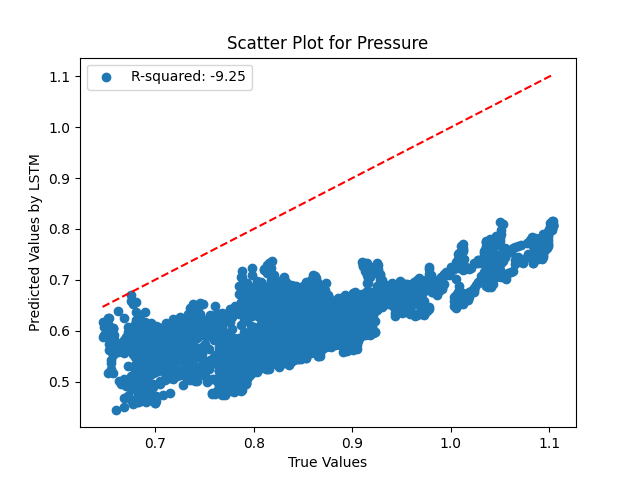}
		\caption{Pressure\\LSTM}
		\label{PressureLSTM}
  \end{minipage}
	\end{figure}

\begin{figure}
		\centering
		\includegraphics[width=\textwidth]{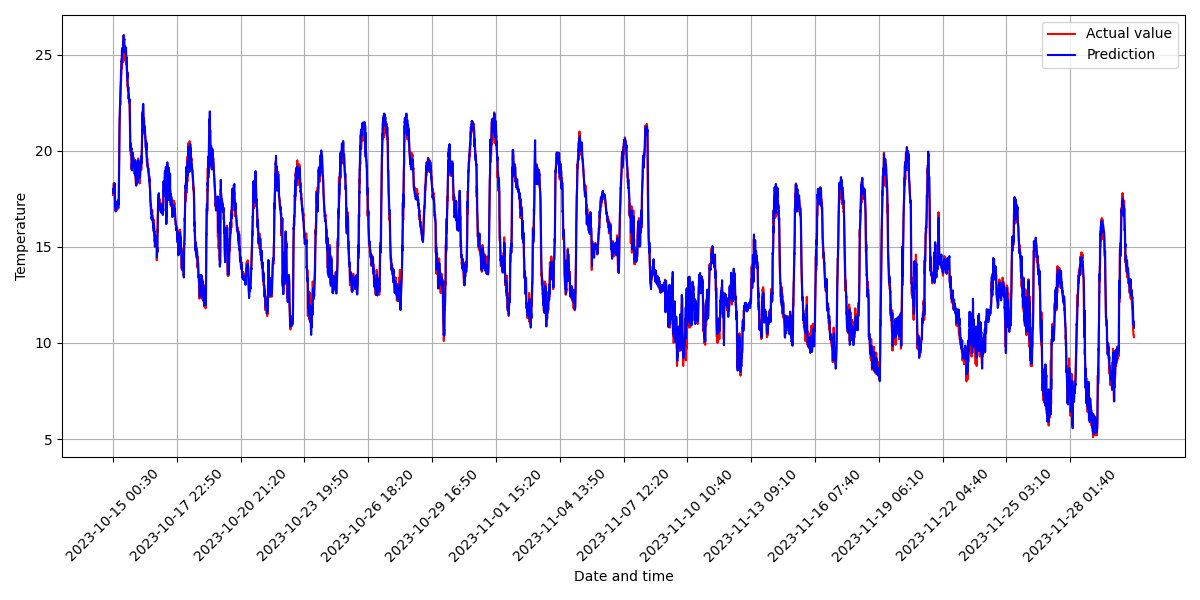}
		\caption{Comparison of the predicted values by our model and the observed values: Temperature}
  \label{fig:comparision1}
\end{figure}

\begin{figure}
		\centering
		\includegraphics[width=\textwidth]{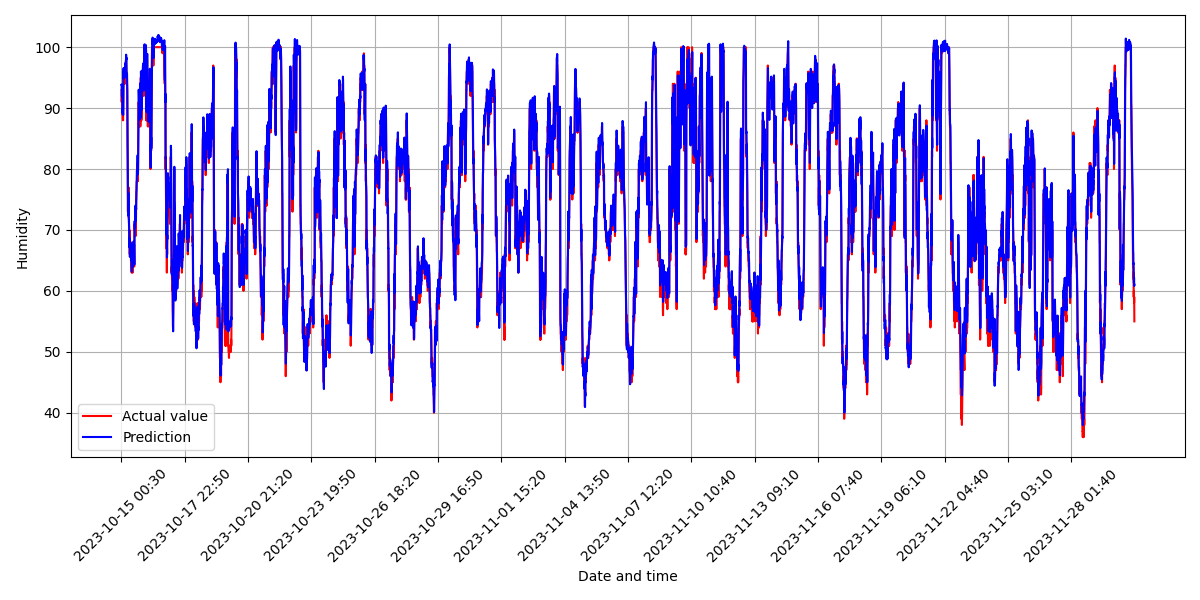}
		\caption{Comparison of the predicted values by our model and the observed values: Humidity}
  \label{fig:comparision2}
	\end{figure}

\begin{figure}
		\centering
		\includegraphics[width=\textwidth]{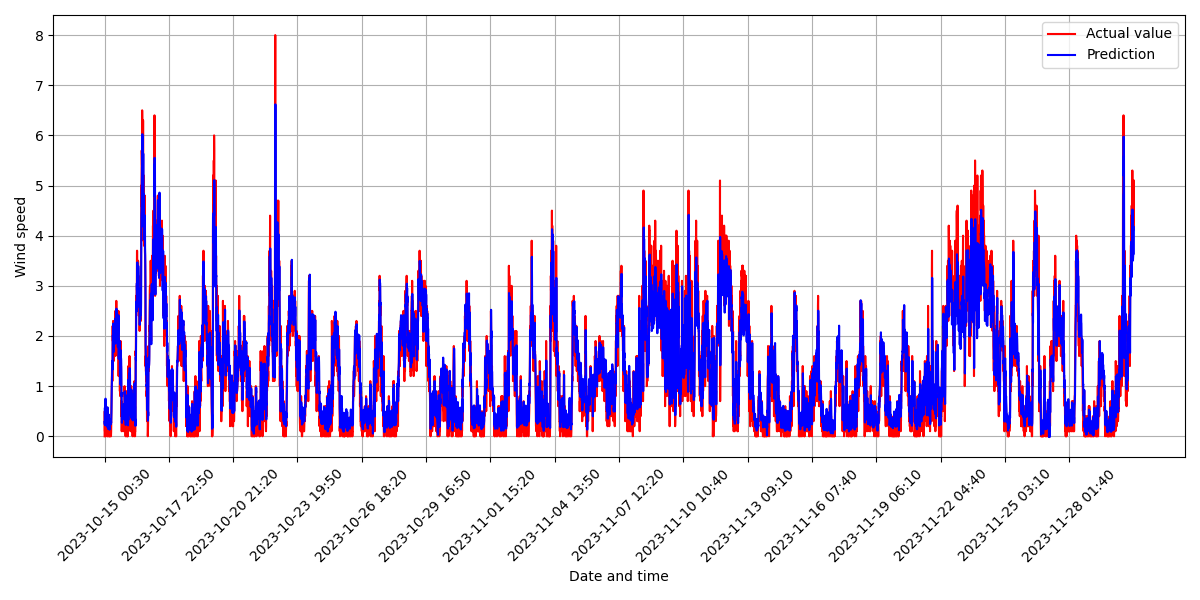}
		\caption{Comparison of the predicted values by our model and the observed values: Wind speed}
  \label{fig:comparision3}
	\end{figure}

\begin{figure}
		\centering
		\includegraphics[width=\textwidth]{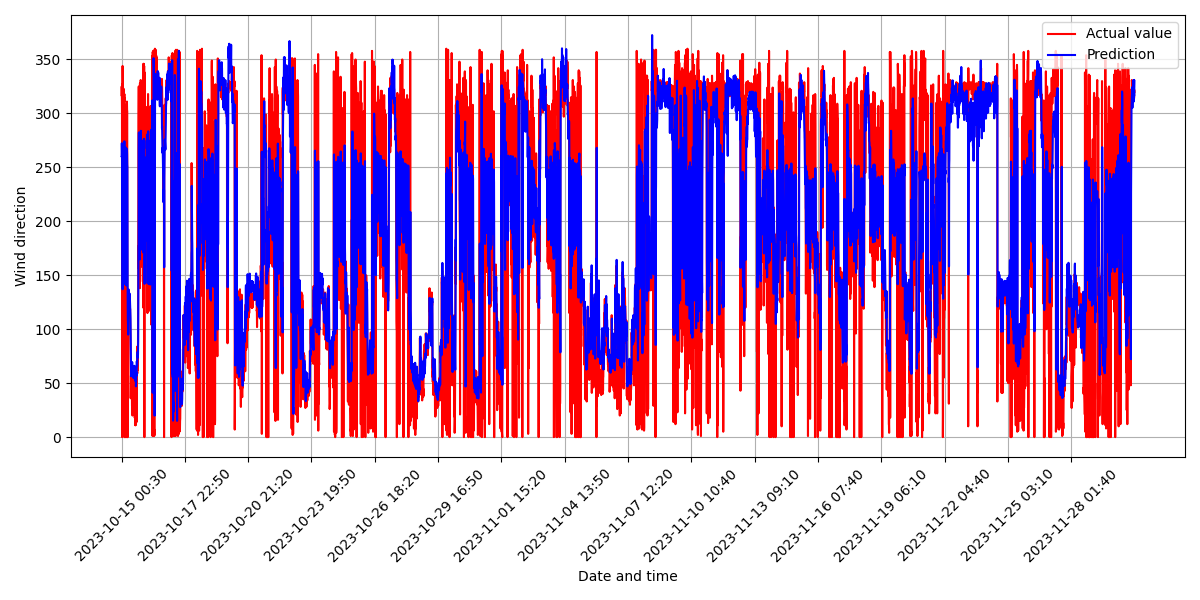}
		\caption{Comparison of the predicted values by our model and the observed values: Wind direction}
  \label{fig:comparision4}
	\end{figure}

\begin{figure}
		\centering
		\includegraphics[width=\textwidth]{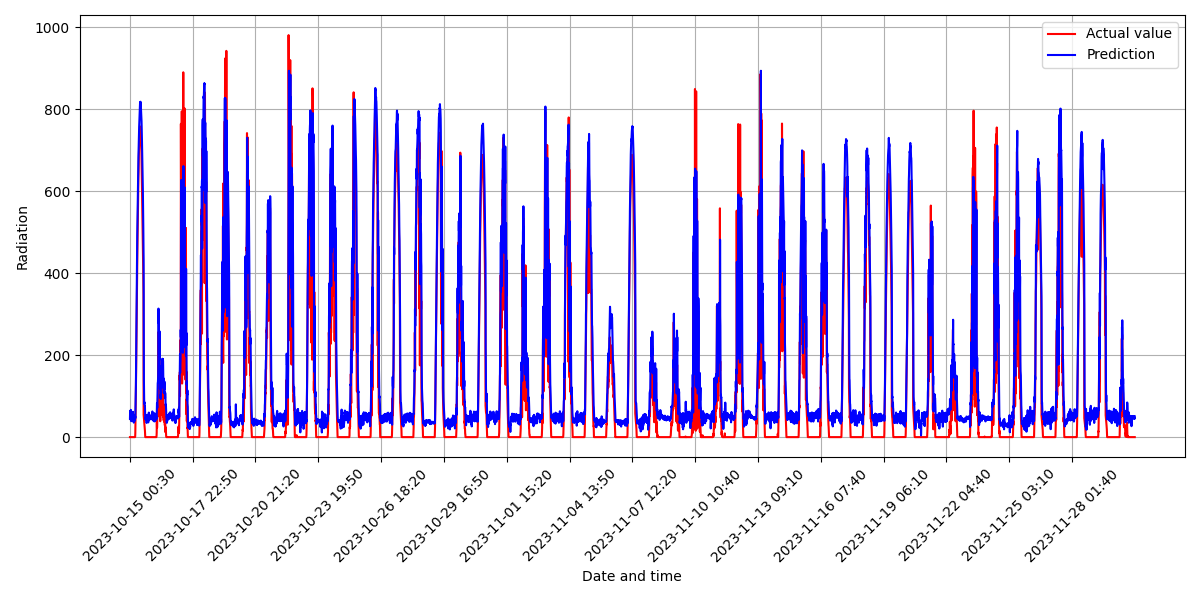}
		\caption{Comparison of the predicted values by our model and the observed values: Radiation}
  \label{fig:comparision5}
\end{figure}

\begin{figure}
		\centering
		\includegraphics[width=\textwidth]{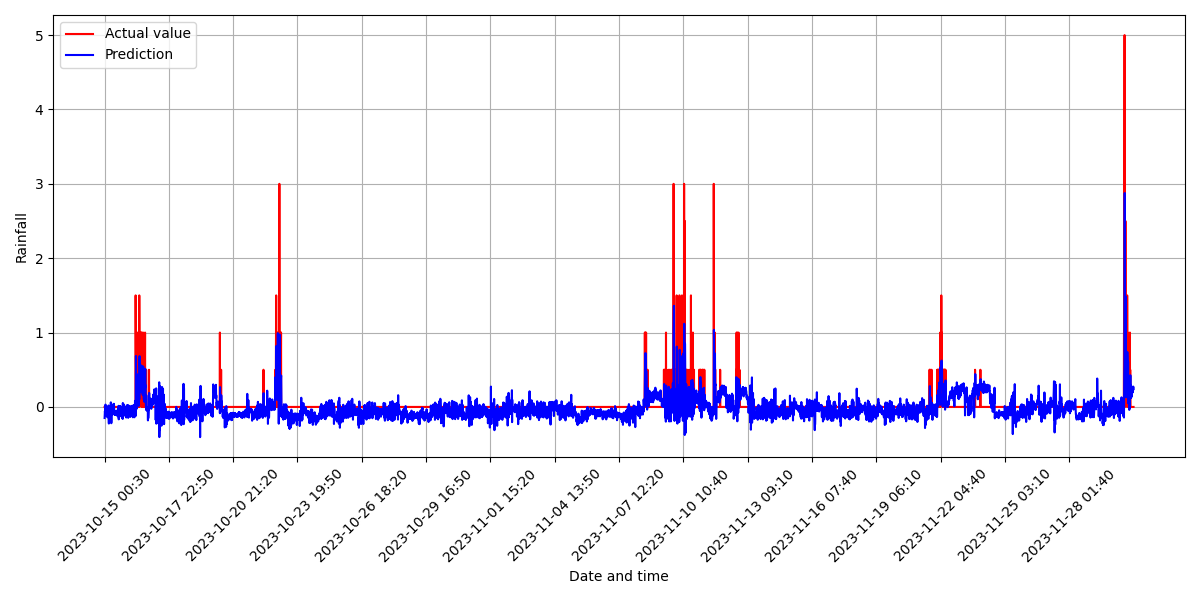}
		\caption{Comparison of the predicted values by our model and the observed values: Rainfall}
  \label{fig:comparision6}
\end{figure}

\begin{figure}
		\centering
		\includegraphics[width=\textwidth]{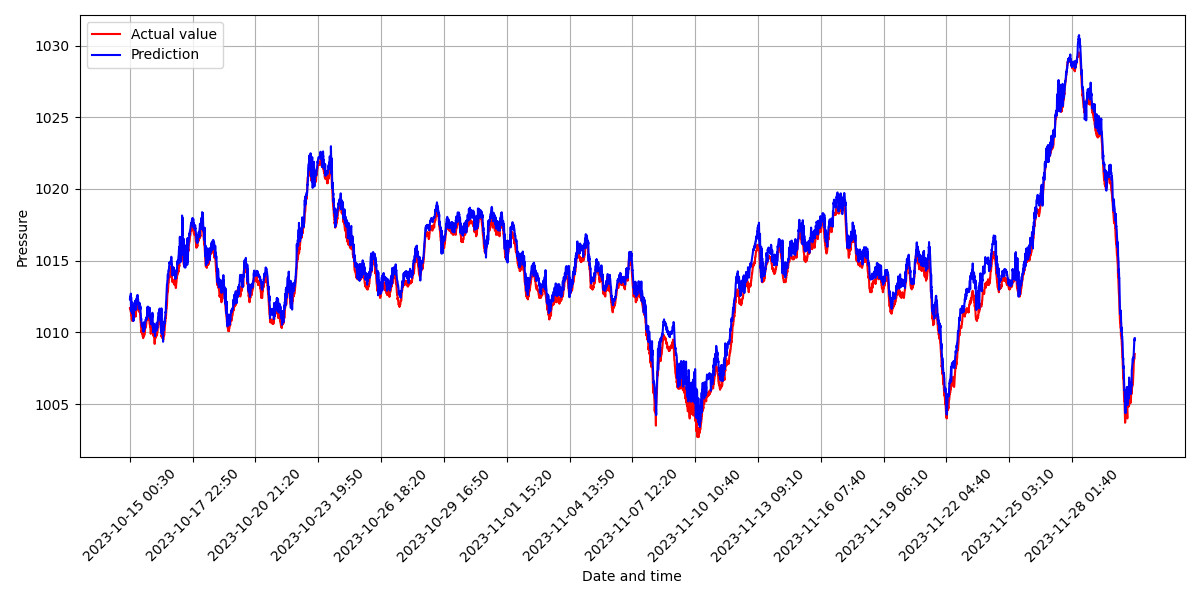}
		\caption{Comparison of the predicted values by our model and the observed values: Pressure}
	\label{fig:comparision7}
\end{figure}

Subsequently, we present the test accuracy of our network, trained over 60 epochs, on the test set. The test loss, which is measured using Mean Squared Error (MSE), Mean Absolute Error (MAE), and Root Mean Squared Error (RMSE), for our model, as well as for Simple RNN and LSTM models, is presented in Table \ref{tab:test loss}. In addition, we assess the correlation between the predicted and true values on the test set for our model, the Simple RNN, and the LSTM model using the Pearson correlation coefficient ($\rho$), as elaborated in Table \ref{tab:correlation}. Furthermore, we report the R-squared coefficient for our model, the Simple RNN, and the LSTM model concerning the predictions and true values in the test set, as shown in Table \ref{tab:r-squared}. For a visual representation, Figures \ref{TemMLP} to \ref{PressureLSTM} display scatter plots for each of the three different models, illustrating the relationships with the weather variables.
To provide a visual assessment for our model, we display a comparison of the predicted values of our model on the test set alongside the observed values in Figures \ref{fig:comparision1}-\ref{fig:comparision7}.

\begin{table}[H]
	\centering
	\caption{Our Network Configuration}
	\begin{tabular}{l c}
		\hline
		\textbf{Parameters} & \textbf{Value} \\ \hline
		Learning rate & 0.7 \\
		Number of hidden nodes & 344 \\
		Batch size & 64 \\ 
		Number of epochs & 60 \\ \hline
	\end{tabular}
	\label{tab:networkconfig}
\end{table}

\begin{table}[H]
	\centering
	\caption{Test loss in MSE, MAE, RMSE}
	\begin{tabular}{l c c c}
		\hline
		\textbf{Models} & \textbf{MSE} & \textbf{MAE} & \textbf{RMSE} \\ \hline
		Our network & 0.007423 & 0.033160 & 0.086158 \\
		Simple RNN & 0.0076136 & 0.035159 & 0.087256 \\
		LSTM & 0.036924 & 0.149351 & 0.192157 \\  \hline
	\end{tabular}
	\label{tab:test loss}
\end{table}


\begin{table}[ht]
	\centering
	\caption{Pearson correlation coefficient ($\rho$) between predicted and true values on the test set}
	\begin{tabular}{cccccccc}
		\hline
		Models & Temperature & Humidity & Wind speed & Wind direction & Radiation  \\
		\hline
		Our network & 0.992 & 0.985 & 0.919 & 0.749 & 0.936  \\
		Simple RNN & 0.987 & 0.981 & 0.914 &  0.737 & 0.932  \\
		LSTM & 0.750 & 0.951 & 0.700 & 0.633 & 0.836 \\
		\hline \\
		\hline
		 &     Rainfall & Pressure \\
		\hline
		  &    0.486 & 0.997 \\
		 &   0.371 & 0.990 \\
		&       0.097 & 0.835\\
		\hline
	\end{tabular}
	\label{tab:correlation}
\end{table}

\begin{table}[ht]
	\centering
	\caption{R-squared between predicted and true values on the test set}
	\begin{tabular}{cccccccc}
		\hline
		Models & Temperature & Humidity & Wind speed & Wind direction & Radiation \\
		\hline
		Our network & 0.966 & 0.968 & 0.831 & 0.551 & 0.870 \\
		Simple RNN & 0.958 & 0.929 & 0.768 & 0.413 & 0.840  \\
		LSTM & 0.409 & 0.641 &  0.175 & 0.228 & 0.523 \\
		\hline
		\hline
		 & Rainfall & Pressure \\
		\hline
		 & 0.018 & 0.993 \\
		 & -0.271 & 0.902 \\
		 & 0.049 & -9.25 \\
		\hline				
	\end{tabular}
	\label{tab:r-squared}
\end{table}

Examining the training loss plot in Figure \ref{Trainingloss}, it is evident that both our model and the Simple RNN consistently achieve lower training losses compared to the LSTM model. We can observe a progressive decrease in loss as the number of training epochs increases, eventually reaching a stable state. This behavior is a direct consequence of employing the minibatch stochastic descent algorithm. Before reaching this stable state, increasing the number of epochs generally correlates with a reduction in training loss. However, once the training loss stabilizes, further fine-tuning of parameters is unlikely to yield substantial improvements. 
Consequently, setting an appropriate number of epochs becomes critical in optimizing the training process efficiently. Conversely, Figure \ref{Trainingaccuracy} demonstrates an inverse trend in training accuracy, where both our model and the Simple RNN consistently achieve higher accuracy compared to the LSTM model. Moreover, when examining Figure \ref{Validationloss} and Figure \ref{Validationaccuracy}, it is evident that throughout the training procedure, both our model and the Simple RNN consistently exhibit lower validation loss and higher accuracy compared to the LSTM model.

In Table \ref{tab:test loss}, our model exhibits the most favorable test loss across all three metrics. Comparing our model and Simple RNN, we observe that they perform remarkably close in terms of test loss, with our model holding a slight advantage. What sets our model apart is its simplicity, resulting in faster training procedures. In contrast, LSTM lags behind both our model and Simple RNN. This discrepancy may be attributed to the inherent complexity of LSTM, which sometimes lead to suboptimal results. Upon examining Table \ref{tab:correlation}, we can observe that our model's predictions are most closely with the observed true data. Simple RNN, while delivering a reasonably accurate performance, falls slightly short of our model. On the other hand, LSTM exhibits the least accurate predictions, suggesting limitations in its ability to model the our dataset effectively. As shown in Table \ref{tab:r-squared}, our model outperforms both the Simple RNN and LSTM models in terms of the R-squared coefficient. However, upon closer examination of the rainfall variable, it becomes evident that all models perform suboptimally. In fact, both the Simple RNN and LSTM models produce negative R-squared values for this variable, indicating poor predictive performance. Regrettably, our model also struggles to predict the rainfall variable effectively.

The low test loss demonstrated above suggests that the proposed model excels in weather prediction within the Itoshima region. Nevertheless, even with a low loss, there remain points for improvement when assessing the network's performance across individual weather condition variables.
As depicted in Figures \ref{TemMLP} to \ref{PressureLSTM}, the proximity of the scatter points to the red reference line serves as an indicator of prediction quality. In the first column of plots, representing Our model, we observe the most favorable performance across all models, particularly evident in variables such as temperature, humidity, and pressure. In the case of wind speed, wind direction, and radiation, both our model and the Simple RNN exhibit superior predictive capabilities when compared to the LSTM model. However, it is important to note that for the rainfall variable, none of the three models achieve satisfactory predictive performance.

Figures \ref{fig:comparision1} to \ref{fig:comparision7} show a comparison of predicted values and true observed values for the seven weather condition variables, revealing the following observations:
\begin{itemize}
	\item The network exhibits excellent performance in capturing the non-linear behavior of temperature, humidity, and pressure. 
	\item Concerning wind speed and wind direction, the network seems to capture the average characteristics of these variables but encounters difficulties with extreme values. This challenge arises due to the high frequency and substantial size of changes in wind direction, making it a formidable task for the network to learn the nuances of these fluctuations.
	\item In the case of radiation, the network generally predicts values accurately but occasionally generates negative values. This anomaly occurs because the MLP network approximates the original function, and radiation data contains numerous zero values. When the network attempts to approximate this zero information in radiation, it occasionally produces negative values, despite their lack of physical meaning in radiation.
	\item The network's performance in rainfall prediction falls short, with predicted values predominantly hovering around zero, occasionally dipping into negative territory. The inherent non-linear nature of the rainfall variable, which comprises a plethora of zero values, poses a challenge for the network in capturing the true essence of rainfall patterns.
\end{itemize}

Based on these observations, it is evident that there is room for improvement in our model's performance. Enhancing its capabilities could involve training it on a larger labeled dataset to enhance stability and equip it better for handling complex and extreme weather conditions. Alternatively, adding more layers to the network could be considered, although the simple addition of fully connected layers mirroring the existing hidden layers may not necessarily yield improved results. Exploring the incorporation of layers from differently structured networks could hold the potential for achieving better performance outcomes.
Additionally, it is noteworthy that we have developed a mathematical model based on Markov chains for forecasting the distribution of rainfall (\cite{GaoTon}). While this model does not provide point estimates for rainfall, it may prove beneficial when integrating additional layers into the network.

\section{Conclusion}

In this study, our primary objective was to present a deep neural network-based methodology for weather forecasting in the Itoshima region. We employed a Multilayer Perceptron (MLP) network comprising three layers with 344 hidden nodes, trained and tested on a meticulously collected dataset specific to the Itoshima region.

Prior to commencing the training phase, several essential preparatory steps were meticulously executed. First, the dataset underwent division into distinct training and test sets, following which a crucial data normalization step was performed, leveraging the min-max scaler method. The determination of hyperparameters, such as learning rate and the number of hidden nodes, was conducted with precision through a model selection process employing 5-fold cross-validation.

Upon rigorous training and meticulous testing, the results yielded by our proposed MLP network were highly promising. The model exhibited exceptional proficiency in predicting fundamental weather variables, including temperature, humidity, and radiation. These predictions consistently aligned closely with observed values, showcasing the model's robust performance.

While our model demonstrated good overall accuracy, we can see its variability in certain weather variables, notably wind speed, wind direction, and rainfall. This variability may come from the dataset's imbalance regarding rainfall and wind. Specifically, numerous values approach zero, while non-zero values are significantly larger.

Of particular concern is the challenge posed by the rainfall variable. During testing, the model encountered difficulties in capturing the fluctuation features of this variable. Addressing this issue will be imperative for enhancing the model's performance in predicting rainfall accurately.
This can be accomplished through the expansion of our dataset and the integration of our mathematical Markov chain model into the network.

\subsubsection*{Funding}
Yuzhong Cheng and Ton Viet Ta were supported by WISE program (MEXT) at Kyushu University.

\bibliographystyle{plain}
\bibliography{base_weather}

\end{document}